\documentclass[11pt]{wlscirep}
\usepackage[utf8]{inputenc}
\usepackage{float}
\usepackage[T1]{fontenc}
\usepackage[draft]{changes}
\usepackage{enumitem}
\usepackage{comment}
\newcommand{\citep}{\cite} %

\usepackage{tabularx}
\usepackage{multirow}
\usepackage{booktabs}
\usepackage{tabulary}  %

\usepackage{listings}
\usepackage{xcolor}

\usepackage{chngcntr}
\usepackage{caption}

\DeclareCaptionType[fileext=lof,placement={!ht},within=none]{suppfigure}[Figure][List of Supplementary Figures]
\DeclareCaptionType[fileext=lot,placement={!ht},within=none]{supptable}[Table][List of Supplementary Tables]

\usepackage{hyperref}
\usepackage{nameref}

\definecolor{codegreen}{rgb}{0,0.6,0}
\definecolor{codegray}{rgb}{0.5,0.5,0.5}
\definecolor{codepurple}{rgb}{0.58,0,0.82}
\definecolor{backcolour}{rgb}{0.95,0.95,0.92}

\lstdefinestyle{mystyle}{
    backgroundcolor=\color{backcolour},
    commentstyle=\color{codegreen},
    keywordstyle=\color{magenta},
    stringstyle=\color{codepurple},
    basicstyle=\ttfamily\footnotesize,
    breakatwhitespace=false,         
    breaklines=true,                 
    keepspaces=true,                
    showspaces=false,                
    showstringspaces=false,
    showtabs=false,                  
    tabsize=2
}

\usepackage{xcolor}         %
\usepackage{tcolorbox}
\usepackage{caption}
\usepackage{wrapfig}
\usepackage{enumitem}
\usepackage{changepage}
\usepackage{subcaption}

\usepackage{colortbl} %
\definecolor{lightgray}{gray}{0.9}
\usepackage{threeparttable} %

\newtcolorbox{findingbox}{
  colback=white,
  colframe=blue!50!black,
  fonttitle=\bfseries,
  title={Finding},
  sharp corners,
  boxrule=1pt,
  left=0pt
}

\usepackage{xcolor}
\definecolor{stanfordred}{RGB}{140,21,21} %
\definecolor{weixingreen}{RGB}{0,150,0}   %
\definecolor{yuhuiblue}{RGB}{0,0,150}     %
\definecolor{binglupurple}{RGB}{128,0,128} %
\definecolor{yianorange}{RGB}{255,140,0}

\title{
Can large language models provide useful feedback on research papers? A large-scale empirical analysis.
}

\author[1*]{Weixin Liang}
\author[1*]{Yuhui Zhang} 
\author[1*]{Hancheng Cao} 
\author[2]{Binglu Wang}
\author[3]{Daisy Yi Ding} 
\author[4]{Xinyu Yang} 
\author[5]{Kailas Vodrahalli} 
\author[3]{Siyu He} 
\author[6]{Daniel Scott Smith} 
\author[4]{Yian Yin} 
\author[6]{Daniel A. McFarland} 
\author[1,3,5+]{James Zou} 

\affil[1]{Department of Computer Science, Stanford University, Stanford, CA 94305, USA}

\affil[2]{Kellogg School of Management, Northwestern University, Evanston, IL 60208, USA}

\affil[3]{Department of Biomedical Data Science, Stanford University, Stanford, CA 94305, USA}

\affil[4]{Department of Information Science, Cornell University, Ithaca, NY 14850, USA}

\affil[5]{Department of Electrical Engineering, Stanford University, Stanford, CA 94305, USA}

\affil[6]{Graduate School of Education, Stanford University, Stanford, CA 94305, USA}

\affil[+]{Correspondence should be addressed to: jamesz@stanford.edu}
\affil[*]{these authors contributed equally to this work}

\begin{abstract}

Expert feedback lays the foundation of rigorous research. However, the rapid growth of scholarly production and intricate knowledge specialization challenge the conventional scientific feedback mechanisms. High-quality peer reviews are increasingly difficult to obtain. Researchers who are more junior or from under-resourced settings have especially hard times getting timely feedback. With the breakthrough of large language models (LLM) such as GPT-4, there is growing interest in using LLMs to generate scientific feedback on research manuscripts. However, the utility of LLM-generated feedback has not been systematically studied. To address this gap, we created an automated pipeline using GPT-4 to provide comments on the full PDFs of scientific papers. We evaluated the quality of GPT-4's feedback through two large-scale studies. We first quantitatively compared GPT-4's generated feedback with human peer reviewer feedback in 15 \emph{Nature} family journals (3,096 papers in total) and the \emph{ICLR} machine learning conference (1,709 papers). The overlap in the points raised by GPT-4 and by human reviewers (average overlap 30.85\% for \emph{Nature} journals, 39.23\% for \emph{ICLR}) is comparable to the overlap between two human reviewers (average overlap 28.58\% for \emph{Nature} journals, 35.25\% for \emph{ICLR}). The overlap between GPT-4 and human reviewers is larger for the weaker papers (i.e., rejected \emph{ICLR} papers; average overlap 43.80\%). We then conducted a prospective user study with 308 researchers from 110 US institutions in the field of AI and computational biology to understand how researchers perceive feedback generated by our GPT-4 system on their own papers. Overall, more than half (57.4\%) of the users found GPT-4 generated feedback helpful/very helpful and 82.4\% found it more beneficial than feedback from at least some human reviewers. While our findings show that LLM-generated feedback can help researchers, we also identify several limitations. 
For example, GPT-4 tends to focus on certain aspects of scientific feedback (e.g., `add experiments on more datasets'), and often struggles to provide in-depth critique of method design. Together our results suggest that LLM and human feedback can complement each other. While human expert review is and should continue to be the foundation of rigorous scientific process, LLM feedback could benefit researchers, especially when timely expert feedback is not available and in earlier stages of manuscript preparation before peer-review. %

\end{abstract}

\begin{document}
\maketitle
\thispagestyle{empty}

\section*{Introduction}

In the 1940s, Claude Shannon, while at Bell Laboratories, embarked on developing a mathematical framework of information and communication \cite{shannon1948mathematical}. Throughout this pursuit, he was faced with the challenge of naming his novel measure and considered terms such as `information' and `uncertainty'. Shannon shared his work with John von Neumann, who quickly recognized the profound links between Shannon's work and statistical mechanics, and proposed what later anchored modern information theory: `Information Entropy' \cite{tribus1971energy}. Scientific progress often rests on feedback and critique. Effective feedback among peer scientists not only elucidates and promotes the way new discoveries are made, interpreted, and communicated, but also catalyzes the emergence of new scientific paradigms by connecting individual insights, coordinating concurrent lines of thoughts, and stimulating constructive debates and disagreement \cite{kuhn1962structure}.

However, the process of providing timely, comprehensive, and insightful feedback on scientific research is often laborious, resource-intensive, and complex \cite{horbach2018changing}. This complexity is exacerbated by the exponential growth in scholarly publications and the deepening specialization of scientific knowledge \cite{price1963little, jones2009burden}. Traditional avenues, such as peer review and conference discussions, exhibit constraints in scalability, expertise accessibility, and promptness. For instance, it has been estimated that peer review -- one of the most major channels of scientific feedback -- costs over 100M researcher hours and \$2.5B US dollars in a single year \cite{aczel2021billion}. 
Yet at the same time, it has been increasingly challenging to secure enough qualified reviewers who can provide high-quality feedback given the rapid growth in the number of submissions \cite{alberts2008reviewing, bjork2013publishing,lee2013bias, kovanis2016global, shah2022challenges}. For example, the number of submissions to the \emph{ICLR} machine learning conference increased from 960 in 2018 to 4,966 in 2023.

While shortage of high-quality feedback presents a fundamental constraint on the sustainable growth of science overall, it also becomes a source of deepening scientific inequalities. Marginalized researchers, especially those from non-elite institutions or resource-limited regions, often face disproportionate challenges in accessing valuable feedback, perpetuating a cycle of systemic scientific inequality \cite{bourdieu2018cultural, merton1968matthew}. 

Given these challenges, there is an urgent need for crafting scalable and efficient feedback mechanisms that can enrich and streamline the scientific feedback process. Adopting such advancements holds the promise of not just elevating the quality and scope of scientific research, given the concerning deceleration in scientific advancements \cite{chu2021slowed, bloom2020ideas}, but also of democratizing its access across the scientific community.

Large language models (LLMs) \cite{brown2020language,ouyang2022training, gpt4}, especially those powered by Transformer-based architectures and pre-trained at immense scales, have opened up great potential in various applications~\citep{ChatGPT-Responses-to-Patient-Questions,lee2022evaluating,chatgpt-pass-medical-exam,chatgpt-wharton-mba}. 
While LLMs have made remarkable strides in various domains, the promises and perils of leveraging LLMs for scientific feedback remain largely unknown.
Despite recent attempts that explore the potential uses of such tools in areas such as automating paper screening \cite{schulz2022future},
error identification \cite{liu2023reviewergpt}, and checklist verification \cite{robertson2023gpt4} \footnote{For a more detailed literature review, see supplementary information.}, we lack large-scale empirical evidence on whether and how LLMs may be used to facilitate scientific feedback and augment current academic practices.

\begin{figure*}[t!] 
\centering
\includegraphics[width=0.75\textwidth]{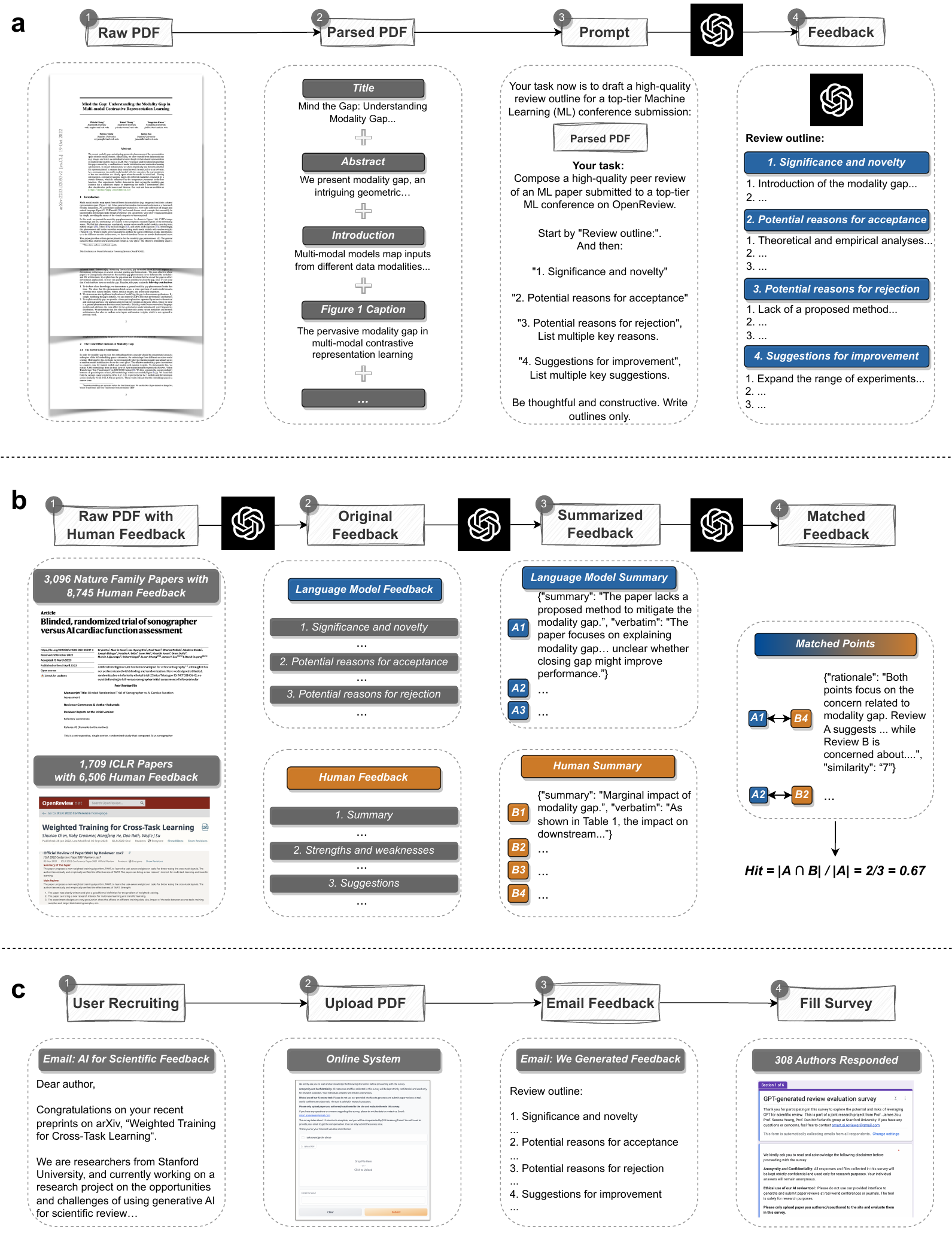}
\caption{
\textbf{Characterizing the capability of LLM in providing helpful feedback to researchers.}
\textbf{a}, 
Pipeline for generating LLM scientific feedback using GPT-4. Given a PDF, we parse and extract the paper's title, abstract, figure and table captions, and main text to construct the prompt. 
We then prompt GPT-4 to provide structured comments with four sections, following the feedback structure of leading interdisciplinary journals and conferences: 
significance and novelty, 
potential reasons for acceptance, 
potential reasons for rejection, and 
suggestions for improvement.
\textbf{b}, 
Retrospective analysis of LLM feedback on 3,096 \emph{Nature} family papers and 1,709 \emph{ICLR} papers. 
We systematically compare LLM feedback with human feedback using a two-stage comment matching pipeline. 
The pipeline first performs extractive text summarization to extract the points of comments raised in LLM and human-written feedback respectively, and then performs semantic text matching to match the points of shared comments between LLM and human feedback. 
\textbf{c}, Prospective user study survey with 308 researchers from 110 US institutions in the field of AI and computational biology. 
Each researcher uploaded a paper they authored, and filled out a survey on the LLM feedback generated for them.  
}
\label{fig:input_output}
\end{figure*}

In this work, we present the first large-scale systematic analysis characterizing the potential reliability and credibility of leveraging LLM for generating scientific feedback. 
Specifically, we developed a GPT-4 based scientific feedback generation pipeline that takes the raw PDF of a paper and produces structured feedback  (Fig.~\ref{fig:input_output}a). 
The system is designed to generate constructive feedback across various key aspects, mirroring the review structure of leading interdisciplinary journals~\cite{nature-for-referees,ncomms-for-reviewers} and conferences~\cite{rogers2021aclrolling,acl23-peer-review-policies,aclijcnlp2021-instructions-reviewers,icml2023-reviewer-tutorial,nicholas2011quick}, including:  
1) Significance and novelty, 
2) Potential reasons for acceptance, 
3) Potential reasons for rejection, and 
4) Suggestions for improvement.

To characterize the informativeness of GPT-4 generated feedback, we conducted both a retrospective analysis and a prospective user study.
In the retrospective analysis, we applied our pipeline on papers that had previously been assessed by human reviewers. We then compared the LLM feedback with the human feedback.
We assessed the degree of overlap between key points raised by both sources to gauge the effectiveness and reliability of LLM feedback. 
Furthermore, we compared the topic distributions of LLM feedback and human feedback. 
To enable such analysis, we curated two complementary datasets containing full-text of papers, their meta information, and associated peer reviews after 2022 \footnote{Focusing data after 2022 avoids bias introduced by 'testing on the training set', since GPT-4, the LLM we used, is trained on data up to Sep 2021 \cite{gpt4}.}. The first dataset was sourced from \emph{Nature} family journals, which are leading scientific journals covering multidisciplinary fields including biomedicine and basic sciences. Our second dataset was sourced from \emph{ICLR} (International Conference on Learning Representations), a leading computer science venue on artificial intelligence. This dataset, although narrower in scope, includes complete reviews for both accepted and rejected papers. These two datasets allowed us to evaluate the performance of LLM in generating scientific feedback across different types of scientific writing (e.g. across fields).

For the prospective user study, we developed a survey in which researchers were invited to evaluate the quality of the feedback produced by our GPT-4 system on their authored papers. 
By analyzing researchers' perspectives on the helpfulness, reliability, and potential limitations of LLM feedback, we can gauge the acceptability and utility of the proposed approach in the manuscript improvement process, and understand stakeholder's subjective perceptions of the framework. 
Through recruitment over institute mailing lists, and contacting paper authors who put preprints on arXiv, we were able to collect survey responses from 308 researchers from 110 US institutions in the field of AI and computational biology that come from diverse education status, experience, and institutes.

\section*{Results}

\subsection*{Generating Scientific Feedback using LLM}

We developed an automated pipeline that utilizes OpenAI's GPT-4~\cite{gpt4} to generate feedback on the full PDF of scientific papers. The pipeline first parses the entire paper from the PDF, then constructs a paper-specific prompt for GPT-4. This prompt is created by concatenating our designed instructions with the paper's title, abstract, figure and table captions, and other main text (\textbf{Fig. \ref{fig:input_output}$a$}, \hyperref[sec:Methods]{Methods}). The prompt is then fed into GPT-4, which generates the scientific feedback in a single pass. Further details and validations of the pipeline can be found in the Supplementary Information.

\subsection*{Retrospective Evaluation}
To evaluate the quality of LLM feedback retrospectively, we systematically assess the content overplay between human feedback given to submitted manuscripts and the LLM feedback using two large-scale datasets. 
The first dataset, sourced from \emph{Nature} family journals, includes 8,745 comments from human reviewers for 3,096 accepted papers across 15 \emph{Nature} family journals, including \emph{Nature}, \emph{Nature Biomedical Engineering}, \emph{Nature Human Behaviour}, and \emph{Nature Communications} (\textbf{Supp. Table~\ref{tab:nature_data}}, \hyperref[sec:Methods]{Methods}). 
The second dataset comprises 6,505 comments from human reviewers for 1,709 papers from the International Conference on Learning Representations (\emph{ICLR}), a leading venue for artificial intelligence research in computer science (\textbf{Supp. Table~\ref{tab:iclr_data}}, \hyperref[sec:Methods]{Methods}). 
These two datasets complement each other: the first dataset (\emph{Nature} portfolio journals) spans a broad range of prominent journals across various scientific disciplines and impact levels, thereby capturing both the universality and variations in human-based scientific feedback. The second dataset (\emph{ICLR}) provides an in-depth perspective of scientific feedback within leading venues of a rapidly evolving field -- machine learning. Importantly, this second dataset includes expert feedback on both accepted and rejected papers.

We developed a retrospective comment matching pipeline to evaluate the overlap between feedback from LLM and human reviewers (\textbf{Fig. \ref{fig:input_output}$b$}, \hyperref[sec:Methods]{Methods}). 
The pipeline first performs extractive text summarization~\cite{luhn1958,edmundson1969new,mihalcea2004textrank,erkan2004lexrank} to extract the comments from both LLM and human-written feedback.
It then applies semantic text matching~\cite{deerwester1990indexing,socher2011dynamic,bowman2015large} to identify shared comments between the two feedback sources. 
We validated the pipeline's accuracy through human verification, yielding an F1 score of 96.8\% for extraction (\textbf{Supp. Table~\ref{tab:combined_verification}$a$}, \hyperref[sec:Methods]{Methods}) and 82.4\% for matching (\textbf{Supp. Table~\ref{tab:combined_verification}$b$}, \hyperref[sec:Methods]{Methods}).

\subsubsection*{LLM feedback significantly overlaps with human-generated feedback}

\begin{figure*}[htb]  
\centering
\includegraphics[width=1\textwidth]{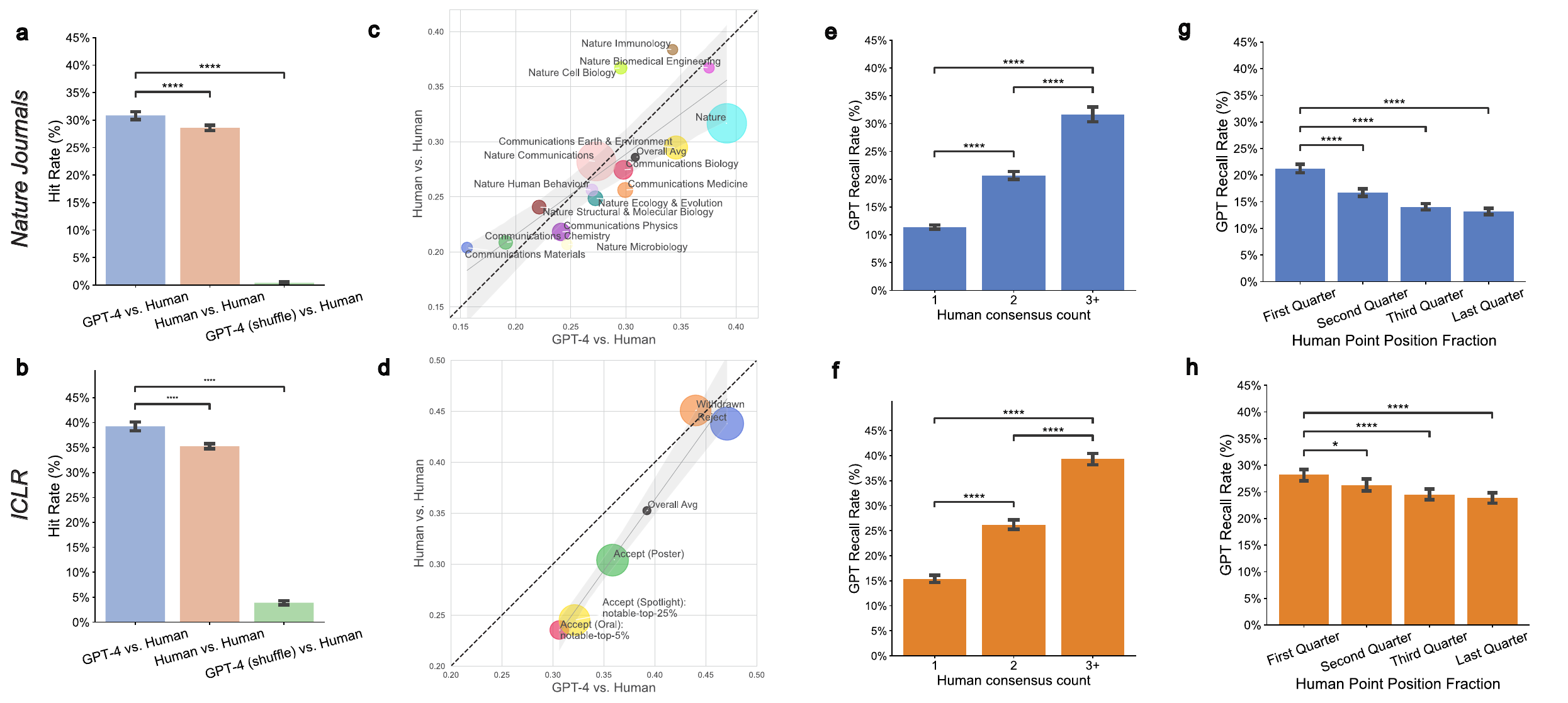}
\caption{
\textbf{Retrospective analysis of LLM and human scientific feedback.}
\textbf{a}, Retrospective overlap analysis between feedback from the LLM versus individual human reviewers on papers submitted to \emph{Nature} Family Journals. 
Approximately one third (30.85\%) of GPT-4 raised comments overlap with the comments from an individual reviewer (hit rate). 
“GPT-4 (shuffle)” indicates feedback from GPT-4 for another randomly chosen paper from the same journal and category. 
As a null model, if LLM mostly produces generic feedback applicable to many papers, 
then there would be little drop in the pairwise overlap between LLM feedback and the comments from each individual reviewer after the shuffling. In contrast, the hit rate drops substantially from 57.55\% to 1.13\% after shuffling, indicating that the LLM feedback is paper-specific. 
\textbf{b}, 
In the International Conference on Learning Representations (\emph{ICLR}), more than one third (39.23\%) of GPT-4 raised comments overlap with the comments from an individual reviewer. 
The shuffling experiment shows a similar result, indicating that the LLM feedback is paper-specific. 
\textbf{c-d}, The overlap between LLM feedback and human feedback appears comparable to the overlap observed between two human reviewers across \emph{Nature} family journals (\textbf{c}) (\(r = 0.80\), \(P = 3.69 \times 10^{-4}\)) and across \emph{ICLR} decision outcomes (\textbf{d}) (\(r = 0.98\), \(P = 3.28 \times 10^{-3}\)). 
\textbf{e-f},  
Comments raised by multiple human reviewers are disproportionately more likely to be hit by GPT-4 on \emph{Nature} Family Journals (\textbf{e}) and \emph{ICLR} (\textbf{f}).  
The X-axis indicates the number of reviewers raising the comment. 
The Y-axis indicates the likelihood that a human reviewer comment matches a GPT-4 comment (GPT-4 recall rate).
\textbf{g-h}, 
Comments presented at the beginning of a reviewer's feedback are more likely to be identified by GPT-4 on \emph{Nature} Family Journals (\textbf{g}) and \emph{ICLR} (\textbf{h}). 
The X-axis indicates a comment's position in the sequence of comments raised by the human reviewer.
Error bars represent 95\% confidence intervals. 
*P < 0.05, **P < 0.01, ***P < 0.001, and ****P < 0.0001.
}
\label{fig:Main-AutoEval}
\end{figure*}

We began by examining the overlap between LLM feedback and human feedback on \emph{Nature} family journal data (\textbf{Supp. Table~\ref{tab:nature_data}}). More than half (57.55\%) of the comments raised by GPT-4 were raised by at least one human reviewer (\textbf{Supp. Fig.~\ref{fig:supplementary-global-hit-rates}$a$}). This suggests a considerable overlap between LLM feedback and human feedback, indicating potential accuracy and usefulness of the system. 
When comparing LLM feedback with comments from each individual reviewer, approximately one third (30.85\%) of GPT-4 raised comments overlapped with comments from an individual reviewer (\textbf{Fig.~\ref{fig:Main-AutoEval}$a$}). The degree of overlap between two human reviewers was similar (28.58\%), after controlling for the number of comments (\hyperref[sec:Methods]{Methods}). 
Results were consistent across other overlapping metrics including Szymkiewicz–Simpson overlap coefficient, Jaccard index, Sørensen–Dice coefficient (\textbf{Supp. Fig.~\ref{fig:supplementary-metrics-vertical}}). This indicates that the overlap between LLM feedback and human feedback is comparable to the overlap observed between two human reviewers.
We further stratified these overlap results by academic journals (\textbf{Fig.~\ref{fig:Main-AutoEval}$c$}). While the degree of overlap between LLM feedback and human comments varied across different academic journals within the \emph{Nature} family — from 15.58\% in \emph{Nature} Communications Materials to 39.16\% in \emph{Nature} — the overlap between LLM feedback and human feedback comments largely mirrored the overlap found between two human reviewers. %
The robustness of the finding further indicates that scientific feedback generated from LLM is similar to what researchers could get from peer reviewers.

In parallel experiments, we investigated the comment overlap between LLM feedback and human feedback on \emph{ICLR} papers data (\textbf{Supp. Table~\ref{tab:iclr_data}}), and the results were largely similar. A majority (77.18\%) of the comments raised by GPT-4 were also raised by at least one human reviewer (\textbf{Supp. Fig.~\ref{fig:supplementary-global-hit-rates}$b$}), indicating considerable overlap between LLM feedback and human feedback. 
When comparing LLM feedback with comments from each individual reviewer, more than one third (39.23\%) of GPT-4 raised comments overlapped with comments from an individual reviewer (\textbf{Fig.~\ref{fig:Main-AutoEval}$b$}). The overlap between two human reviewers was similar (35.25\%), after controlling for the number of comments (\hyperref[sec:Methods]{Methods}, \textbf{Supp. Fig.~\ref{fig:supplementary-metrics-vertical}}). 
We further stratified these overlap results by the decision outcomes of the papers (\textbf{Fig.~\ref{fig:Main-AutoEval}$d$}). Similar to results over \emph{Nature} family journals, we found that the overlap between LLM feedback and human feedback comments largely mirrored the overlap found between two human reviewers. %

In addition, as \emph{ICLR} dataset includes both accepted and rejected papers, we conducted stratification analysis and found a correlation between worse acceptance decisions and larger overlap in \emph{ICLR} papers. Specifically, papers accepted with oral presentations (representing the top 5\% of accepted papers) have an average overlap of 30.63\% between LLM feedback and human feedback comments. The average overlap increases to 32.12\% for papers accepted with a spotlight presentation (the top 25\% of accepted papers), while rejected papers bear the highest average overlap at 47.09\%. 
A similar trend was observed in the overlap between two human reviewers: 23.54\% for papers accepted with oral presentations (top 5\% accepted papers), 24.52\% for papers accepted with spotlight presentations (top 25\% accepted papers), and 43.80\% for rejected papers. This suggests that rejected papers may have more apparent issues or flaws that both human reviewers and LLMs can consistently identify. Additionally, the increased overlap between LLM feedback and actual human reviewer feedback for rejected papers indicates that LLM feedback could be particularly constructive and formative for papers that require more substantial revisions to be accepted. Indeed, by raising these concerns earlier in the scientific process before review, these papers and the science they report may be improved.

\subsubsection*{LLM could generate non-generic feedbacks.}

Is it possible that LLM merely generates generic feedback applicable to multiple papers? A potential null model is that LLM mostly produces generic feedback applicable to many papers. To test this hypothesis, we performed a shuffling experiment aimed at verifying the specificity and relevance of LLM generated feedback. 
For each paper in the \emph{Nature} family journal data, the LLM feedback was shuffled for papers from the same journal and within the same \emph{Nature} category (\hyperref[sec:Methods]{Methods}). 
If the LLM were producing only generic feedback, we would observe no decrease in the pairwise overlap between shuffled LLM feedback and human feedback. 
In contrast, the pairwise overlap significantly decreased from 30.85\% to 0.43\% after shuffling (\textbf{Fig.~\ref{fig:Main-AutoEval}$a$}). A similar drop from 39.23\% to 3.91\% was observed on \emph{ICLR} (\textbf{Fig.~\ref{fig:Main-AutoEval}$b$}). These results suggest that LLM feedback is paper-specific.

\subsubsection*{LLM is consistent with humans on major comments}

What characteristics do LLMs' comments exhibit? What are the distinctive features of the human comments that align with LLMs'? Here we evaluate the unique characteristics of comments generated by LLMs. 
Our analysis revealed that comments identified by multiple human reviewers are more likely to be echoed by LLMs. 
For instance, in the \emph{Nature} family journal data (\textbf{Fig.~\ref{fig:Main-AutoEval}$e$}), a comment raised by a single human reviewer had an 11.39\% chance of being identified by LLMs. This probability increased to 20.67\% for comments raised by two reviewers, and further to 31.67\% for comments raised by three or more reviewers. 
A similar trend was observed in the \emph{ICLR} data (\textbf{Fig.~\ref{fig:Main-AutoEval}$f$}), where the likelihood of LLMs identifying a comment increased from 15.39\% for a single reviewer to 26.21\% for two reviewers, and 39.33\% for three or more reviewers. 
These findings suggest that LLMs are more likely to identify common issues or flaws that are consistently recognized by multiple human reviewers, compared to specific comments raised by a single reviewer. 
This alignment of LLM with human perspectives indicates its ability to identify what is generally considered as major or significant issues.

We further examined the likelihood of LLM comments overlapping with human feedback based on their position in the sequence, as earlier comments in human feedback (e.g. ``concern 1'') may represent more significant issues. 
To this end, we divided each human reviewer's comment sequence into four quarters within the Nature journal data (Fig.~\ref{fig:Main-AutoEval}$g$). 
Our findings suggest that comments raised in the first quarter of the review text are most likely (21.23\%) to overlap with LLM comments, with subsequent quarters revealing decreasing likelihoods (16.74\% for the second quarter). Similar trends were observed in the ICLR papers data, where earlier comments in the sequence showed a higher probability of overlap with LLM comments (Fig.~\ref{fig:Main-AutoEval}$h$). These findings further support that LLM tends to align with human perspectives on what is generally considered as major or significant issues.

\subsubsection*{LLM feedback emphasizes certain aspects more than humans}
We next analyzed whether certain aspects of feedback are more/less likely to be raised by the LLM and human reviewers. We focus on \emph{ICLR} for this analysis, as it's more homogeneous than \emph{Nature} family journals, making it easier to categorize the main aspects of review.
Drawing on existing research in peer review literature within the machine learning domain~\cite{values-in-ML-research,smith2022real,koch2021reduced,scheuerman2021datasets}, we developed a schema comprising 11 distinct aspects of comments. We then performed human annotation on a randomly sampled subset (\hyperref[sec:Methods]{Methods}).

\begin{figure*}[htb]  
\centering
\includegraphics[width=0.55\textwidth]{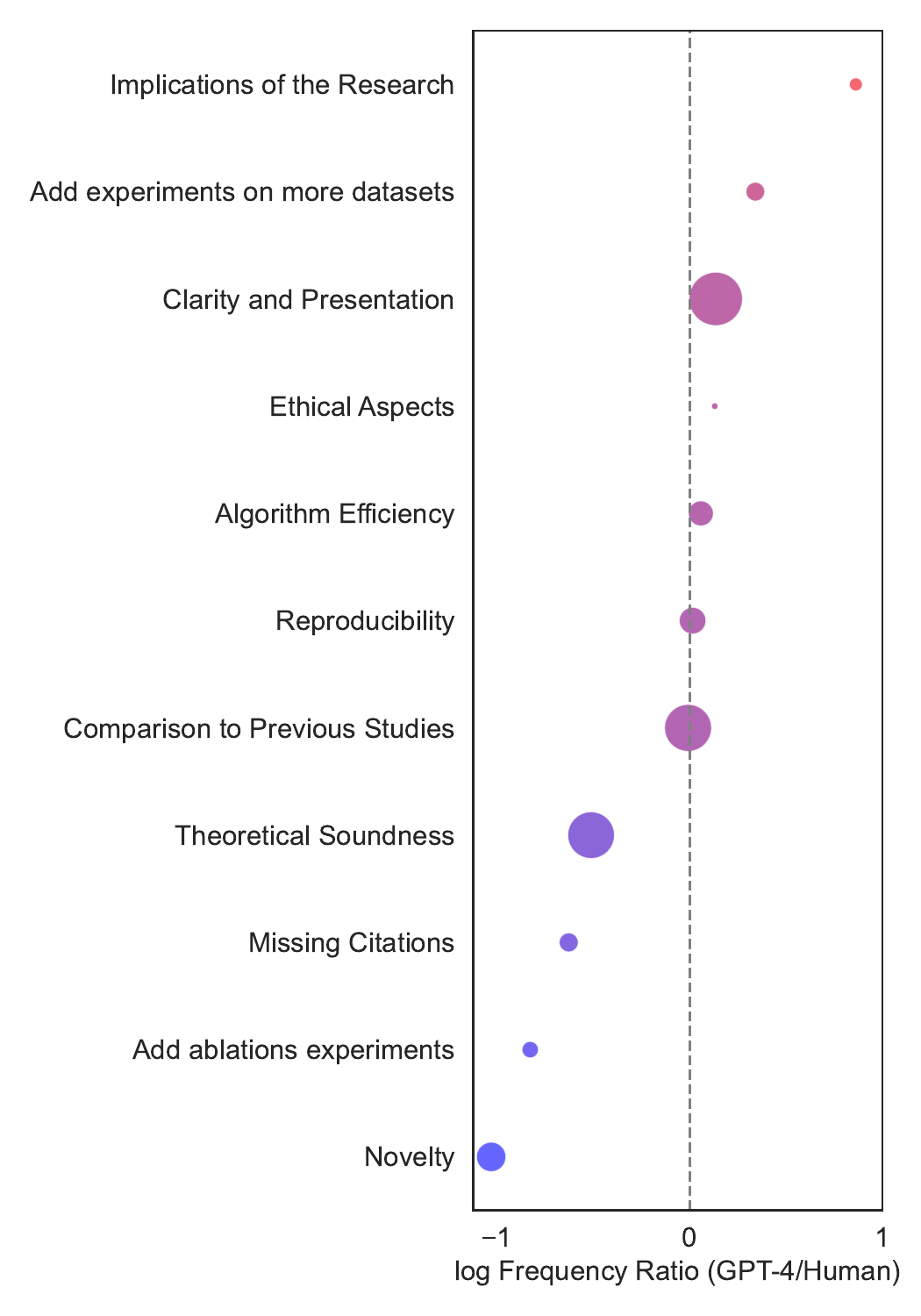}
\caption{
\textbf{LLM based feedback emphasizes certain aspects more than humans.} 
LLM comments on the implications of research 7.27 times more frequently than human reviewers. 
Conversely, LLM is 10.69 times less likely to comment on novelty compared to human reviewers. 
While both LLM and humans often suggest additional experiments, their focuses differ: human reviewers are 6.71 times more likely than LLM to request additional ablation experiments, whereas LLM is 2.19 times more likely than humans to request experiments on more datasets. 
Circle size indicates the prevalence of each aspect in human feedback. 
}
\label{fig:ICLR_value}
\end{figure*}

\textbf{Fig.~\ref{fig:ICLR_value}} presents the relative frequency of each of the 11 aspects of comments raised by humans and LLM. LLM comments on the implications of research 7.27 times more frequently than humans do. Conversely, LLM is 10.69 times less likely to comment on novelty than humans are. While both LLM and humans often suggest additional experiments, their focuses differ: humans are 6.71 times more likely than LLM to request more ablation experiments, whereas LLM is 2.19 times more likely than humans to request experiments on more datasets. These findings suggest that the emphasis put on certain aspects of comments varies between LLMs and human reviewers. This variation highlights the potential advantages that a human-AI collaboration could provide. Rather than having LLM fully automate the scientific feedback process, humans can raise important points that LLM may overlook. Similarly, LLM could supplement human feedback by providing more comprehensive comments.

\subsection*{Prospective User Study and Survey}

Taken together, our retrospective evaluations above suggest that LLMs can generate scientific feedback that focuses on similar aspects as human reviewers. Yet, consistency in concerns is only one of the many factors contributing to the utility of scientific feedback. Recent studies in human-AI interaction have also identified additional factors that individuals may consider when evaluating and adopting AI-based tools \cite{lee2022evaluating}, prompting us to ask: how do scientific researchers respond to feedback generated by LLMs? Thus, we launched a survey study on 308 researchers from 110 US institutions who opted in to receive LLM-generated scientific feedback on their own papers, and were asked to evaluate its utility and performance. While our sampling approach is subject to  biases of self-selection, the data can provide valuable insights and subjective perspectives from researchers  complementing our retrospective analysis \cite{meyer2015household,ross2022women}. The results from the user study are illustrated in \textbf{Fig.~\ref{fig:user_study}}.

Our user study provides additional evidence that is largely consistent with retrospective evaluations. First, the user study survey results corroborate the findings from the retrospective evaluation on significant overlaps between LLM feedback and human feedback: more than 70\% of participants think there is at least “partial alignment” between LLM feedback and what they think/would expect on the significant points and issues with their paper, and 35\% of participants think the alignment is considerable or substantial (\textbf{Fig.~\ref{fig:user_study}$b$}). Second, the survey study further corroborates the findings from the automated evaluation on the ability of the language model to generate non-generic feedback: 32.9\% of participants think our system-generated feedback is “less specific than many, but more specific than some peer reviewers", while 17.3\% and 14\% think it is “about as specific as peer reviewers", or “more specific than many peer reviewers", further corroborating that LLMs can generate non-generic reviews (\textbf{Fig.~\ref{fig:user_study}$d$}).

\subsubsection*{Researchers find LLM feedback helpful}
\begin{figure*}[t!]  
\centering
\includegraphics[width=1\textwidth]{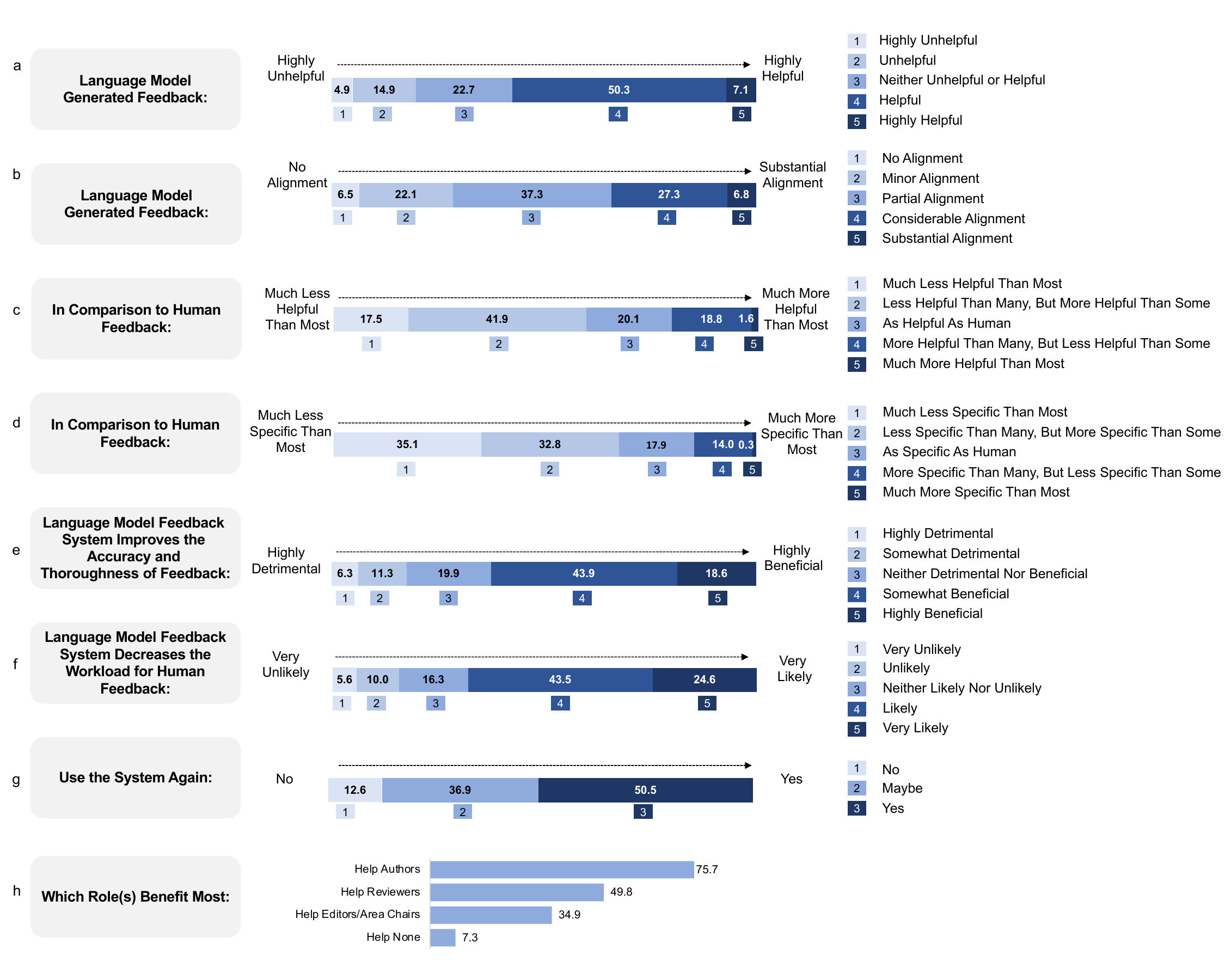}
\caption{
\textbf{Human study of LLM and human review feedback ($n = 308$).}
\textbf{a-b}, LLM generated feedback is generally helpful and has substantial overlaps with actual feedback from human reviewers.
\textbf{c-d}, Compared to human feedback, LLM feedback is slightly less helpful and less specific.
\textbf{e-f}, Users generally believe that the LLM feedback system can improve the accuracy and thoroughness of reviews, and reduce the workload of reviewers.
\textbf{g}, Most users intend to use or potentially use the LLM feedback system again.
\textbf{h}, Users believe that the LLM feedback system mostly helps authors, followed by reviewers and editors / area chairs.
Numbers are percentages (\%).
}
\label{fig:user_study}
\end{figure*}

Participants were surveyed about the extent to which they found the LLM feedback helpful in improving their work or understanding of a subject. The majority responded positively, with over 50.3\% considering the feedback to be helpful, and 7.1\% considering it to be very helpful (\textbf{Fig.~\ref{fig:user_study}$a$}). When compared with human feedback, while 17.5\% of participants considered it to be inferior to human feedback, 41.9\% considered it to be less helpful than many, but more helpful than some human feedback. Additionally, 20.1\% considered it to be about the same level of helpfulness as human feedback, and 20.4\% considered it to be even more helpful than human feedback (\textbf{Fig.~\ref{fig:user_study}$c$}). Our evaluation also revealed that the perceptions of alignment and helpfulness were consistent across various demographic groups. Individuals from different educational backgrounds, ranging from undergraduate to postgraduate levels, found the feedback equally helpful and aligned with human feedback. Similarly, whether an experienced or a novice researcher, participants across the spectrum of publishing and reviewing experience reported similar levels of satisfaction and utility from the LLM based feedback, indicating that LLM based feedback tools could potentially be helpful to a diverse range of population (\textbf{Supp. Fig.~\ref{fig:plots_helpfulness_years_pdf},\ref{fig:plots_helpfulness_status_pdf}}).

In line with the helpfulness of the system, 50.5\% of survey participants further expressed their willingness to reuse the system (\textbf{Fig.~\ref{fig:user_study}$g$}). The participants expressed optimism about the potential improvements that continued use of the system could bring to the traditional human feedback process (\textbf{Fig.~\ref{fig:user_study}$e,f$}). They believe that the LLM technology can further refine the quality of reviews and possibly introduce new capabilities. Interestingly, the evaluation also revealed that participants believe authors are more likely to benefit from LLM based feedback than other stakeholders such as reviewers, and area chairs (\textbf{Fig.~\ref{fig:user_study}$h$}). Many participants envisioned a timely feedback tool for authors to receive comments on their papers in a timely manner, e.g. one participant wrote,
``The review took five minutes and was of a reasonably high quality. This can tremendously help authors to receive a fast turnaround feedback and help in polishing their submissions.'' Another participant wrote,
``After writing a paper or a review, GPT could help me gain another perspective to re-check the paper.''

\subsubsection*{LLM could generate novel feedback not mentioned by humans.}

Beyond generating feedback that aligns with humans, our results also suggest that LLM could potentially generate useful feedback that has not been mentioned by humans, e.g., 65.3\% of participants think at least to some extent LLM feedback offers perspectives that have been overlooked or underemphasized by humans.
Several participants mentioned that:
\begin{itemize}[noitemsep,topsep=0pt]
    \item ``It consists more points, covering aspects which human may forget to think about.''
    \item ``It actually highlighted a few limitations which human reviewers didn't point out to, but as authors we were aware of it and were expecting it. But this GPT figured out some of them, so that's interesting.''
    \item ``The GPT-generated review suggested me to do visualization to make a more concrete case for interpretability. It also asked to address data privacy issues. Both are important, and human reviewers missed this point.''
\end{itemize}

\subsubsection*{Limitations of LLM feedback}
Study participants also discussed limitations of the current system. The most important limitation is its ability to generate specific and actionable feedback, e.g.
\begin{itemize}[noitemsep,topsep=0pt]
    \item ``Potential Reasons are too vague and not domain specific.''
    \item ``GPT cannot provide specific technical areas for improvement, making it potentially difficult to improve the paper.''
    \item ``The reviews crucially lacked much in-depth critique of model architecture and design, something actual reviewers would be able to comment on given their likely considerable experience in fields closely related to the focus of the paper.''
\end{itemize}
As such, one future direction to improve the LLM based scientific feedback system is to nudge the system towards generating more concrete and actionable feedback, e.g. through pointing to specific missing work, experiments to add. As one participant nicely summarized:
\begin{itemize}[noitemsep,topsep=0pt]
    \item ``(large languge model generated) reviews were less about the content and more about the testing regime as well as less ML details-focused, but this is okay as it still gave relevant and actionable advice on areas of improvement in terms of paper layout and presenting results. GPT-generated reviews are especially useful here when less-experience authors may leave out details on implementation and construction or forget to thoroughly explain testing regime by providing pointers on areas to polish the paper in, potentially decreasing the number of review cycles before publication.''
\end{itemize}

\section*{Discussion}

In this study, we characterized the usefulness and reliability of LLM in scientific evaluation by building and evaluating an LLM-based scientific feedback generation framework. Through a combination of retrospective (comparing LLM feedback with human feedback from the peer review process) and prospective evaluation design (user study with researchers), we have seen a substantial level of overlap and positive user perceptions regarding the usefulness of LLM feedback. Furthermore, in evaluating user perceptions, we found that a majority of participants regarded LLM feedback as useful in the manuscript improvement process, and sometimes LLM could bring up novel points not covered by humans. The positive feedback from users highlights the potential value and utility of leveraging LLM feedback as a valuable resource for authors seeking constructive feedback and suggestions for enhancing their manuscripts. This could be especially helpful for researchers who lack access to timely quality feedback mechanisms, e.g., researchers from traditionally underprivileged regions who may not have resources to access conferences, or even peer review (their works are much more likely than those of “mainstream” researchers to get desk rejected by journals and thus seldom go through the peer review process \cite{merton1968matthew}). For others, the framework could be used as a  mechanism for authors to self-check and improve their work in a timely manner, especially in an age of exponentially growing scientific papers and increasing challenges to secure timely and quality peer reviewer feedback. Our analysis suggests that people from diverse educational backgrounds and publishing experience can find the LLM scientific feedback generation framework useful (\textbf{Supp. Fig.~\ref{fig:plots_helpfulness_years_pdf},\ref{fig:plots_helpfulness_status_pdf}}).

Despite the potential of LLMs in providing timely and helpful scientific feedback, it is important to note that expert human feedback will still be the cornerstone of rigorous scientific evaluation. As demonstrated in our findings, our analysis reveals limitations of the framework, e.g., LLM is biased towards certain aspects of scientific feedback (e.g., “add experiments on more datasets”), and sometimes feels “generic” to the authors (while participants also indicate that quite often human reviewers are “generic”). While comparable and even better than some reviewers, the current LLM feedback cannot substitute specific and thoughtful human feedback by domain experts.

It is also important to note the potential misuse of LLM for scientific feedback. We argue that LLM feedback should be primarily used by researchers identify areas of improvements in their manuscripts prior to official submission. It is important that expert human reviewers should deeply engage with the manuscripts and provide independent assessment without relying on LLM feedback. Automatically generating reviews without thoroughly reading the manuscript would undermine the rigorous evaluation process that forms the bedrock of scientific progress.

More broadly, our study contributes to the recent discussions on the impacts of LLM and generative AI on existing work practices. Researchers have discussed the potential of LLM to improve productivity \cite{noy2023experimental,peng2023impact}, creativity \cite{epstein2023art}, and facilitate scientific discovery \cite{wang2023scientific}. We envision that LLM and generative AI, if deployed responsibly, could also potentially bring a paradigm change to how researchers conduct research, collaborate, and provide evaluations, influencing the way science and technology advance. Our work brings a preliminary investigation into such potentials through a concrete prototype for scientific feedback.

There are several limitations to our study that are important to highlight. First, our results are based on one specific instantiation of scientific feedback from LLM, i.e., our framework is based on the GPT-4 model, enabled by a specific prompt. While we have spent significant efforts in improving the performance of our GPT-4 feedback pipeline (and achieved reasonable utility), the results should be interpreted as a lower bound, rather than an upper bound, on the potential of leveraging LLMs for scientific feedback. Moreover, our system only leverages zero-shot learning of GPT-4 without fine-tuning on additional datasets. Further, the architecture and prompt used in our study only represent one of the many possible forms of LLM-based scientific feedback. Aside from exploring other LLMs and conducting more sophisticated prompt engineering, future work could incorporate labeled datasets of “high quality scientific feedback” to further fine-tune the LLM, or prompt LLM to leverage tools (e.g., fact check API, algorithm analysis API) so that the feedback could be more detailed and method-specific. Nevertheless, our proposed framework proves to be helpful and aligns well with comments brought up by human reviewers, demonstrating the  potential of incorporating LLM in the scientific evaluation process, echoing prior works arguing for AI in the scientific process \cite{wang2023scientific}.  Our retrospective evaluation used \emph{Nature} family data and \emph{ICLR} data. While these data cover a range of scientific fields, including biology, computer science, etc., and the \emph{ICLR} dataset includes both accepted and rejected papers, all studied papers are targeted at top venues in English. Future work should further evaluate the framework with more coverage. Our user study is limited in coverage of participant population and suffer from a self-selection issue. Current results are based on responses from researchers in machine learning and computational biology, and while we aim to ensure representativeness of researchers by reaching out to randomly sampled authors who upload preprints to arXiv in recent months, participants who opt into the study are likely to be interested and familiar with LLM or AI in general. Finally, the current version of the GPT-4 model we utilized does not possess the capability to understand or interpret visual data such as tables, graphs, and figures, which are integral components of scientific literature. Future iterations could explore integrating visual LLMs or specialized modules that can comprehend and critique visual elements, thereby offering a more comprehensive form of scientific feedback.

One direction for future work is to explore the extent to which the proposed approach can help identify and correct errors in scientific papers. This would involve artificially introducing various types of errors, including typos, mistakes in data analyses, and errors in mathematical equations. By evaluating whether LLM-generated feedback can effectively detect and rectify these errors, we can gain further insights into the system's ability to improve the overall accuracy and quality of scientific manuscripts. Furthermore, investigating the limitations and challenges associated with error detection and correction by LLM is crucial. This includes understanding the types of errors that may be more challenging for the model to detect and correct, as well as evaluating the potential impact of false positives or false negatives in the generated feedback. Such insights can inform the development of more robust and accurate AI-assisted review systems. Additionally, we intend to broaden the scope of evaluated scientific papers to include manuscripts written in languages other than English, or by authors for whom English is not their first language, in order to understand whether LLMs can provide useful feedback for such papers.

\section*{Methods}
\label{sec:Methods}

\subsection*{Overview of the \emph{Nature} Family Journals Dataset}

Several journals within the \emph{Nature} group have adopted a transparent peer review policy, enabling authors to publish reviewers' comments alongside the accepted papers~\cite{nature2020-transparent-review}. For instance, by 2021, approximately 46\% of \emph{Nature} authors opted to make their reviewer discussions public \cite{nature-trialling-review}. Our dataset comprises papers from 15 \emph{Nature} family journals, published between January 1, 2022, and June 17, 2023. 
We sourced papers from 15 \emph{Nature} family journals, focusing on those published between January 1, 2022, and June 17, 2023. Within this period, our dataset includes 773 accepted papers from \emph{Nature} with 2,324 reviews, 810 sampled accepted papers from \emph{Nature} Communications with 2,250 reviews, and many others. 
In total, our dataset includes 3,096 accepted papers and 8,745 reviews (\textbf{Supp. Table~\ref{tab:nature_data}}). The data were sourced directly from the \emph{Nature} website (\url{https://nature.com/}).

\subsection*{Overview of the \emph{ICLR} Dataset}

The International Conference on Learning Representations (\emph{ICLR}) is a leading publication venue in the machine learning field. \emph{ICLR} implements an open review policy, making reviews for all papers accessible, including those for rejected papers. Accepted papers at \emph{ICLR} are categorized into Oral presentations (top 5\% of papers), Spotlight (top 25\%), and Poster presentations. In 2022, \emph{ICLR} received 3,407 submissions, which increased to 4,966 by 2023. Using a stratified sampling method, we included 55 Oral (with 200 reviews), 173 Spotlight (664 reviews), 197 Poster (752 reviews), 213 rejected (842 reviews), and 182 withdrawn (710 reviews) papers from 2022. For 2023, we included 90 Oral (317 reviews), 200 Spotlight (758 reviews), 200 Poster (760 reviews), 212 rejected (799 reviews), and 187 withdrawn (703 reviews) papers. The dataset comprises 1709 papers and 6,506 reviews in total (\textbf{Supp. Table~\ref{tab:iclr_data}}). The paper PDFs and corresponding reviews were retrieved using the OpenReview API (\url{https://docs.openreview.net/}).

\subsection*{Generating Scientific Feedbacks using LLM}

We prototyped a pipeline to generate scientific feedback using OpenAI's GPT-4~\cite{gpt4} (\textbf{Fig. \ref{fig:input_output}$a$}). The system's input was the academic paper in PDF format, which was then parsed with the machine-learning-based ScienceBeam PDF parser~\cite{ecer2017sciencebeam}. Given the token constraint of GPT-4, which allows 8,192 tokens for combined input and output, the initial 6,500 tokens of the extracted title, abstract, figure and table captions, and main text were utilized to construct the prompt for GPT-4 (\textbf{Supp. Fig.~\ref{fig:supplementary-input_output}}). This token limit exceeds the 5,841.46-token average of \emph{ICLR} papers and covers over half of the 12,444.06-token average for \emph{Nature} family journal papers (\textbf{Supp. Table~\ref{tab:paper_review_length}}).
For clarity and simplicity, we instructed GPT-4 to generate a structured outline of scientific feedback. 
Following the reviewer report instructions from machine learning conferences~\cite{rogers2021aclrolling,acl23-peer-review-policies,aclijcnlp2021-instructions-reviewers,icml2023-reviewer-tutorial,nicholas2011quick} and \emph{Nature} family journals~\cite{nature-for-referees,ncomms-for-reviewers}, we provided specific instructions to generate four feedback sections: significance and novelty, potential reasons for acceptance, potential reasons for rejection, suggestions for improvement (\textbf{Supp. Fig.~\ref{fig:Nature-prompt-generate}}). 
The feedback for each paper was generated by GPT-4 in a single pass.

\subsection*{Retrospective Extraction and Matching of Comments from Scientific Feedback}

To evaluate the overlap between LLM feedback and human feedback, we developed a two-stage comment matching pipeline (\textbf{Supp. Fig.~\ref{fig:supplementary-retrospective_evaluation}}). 
In the first stage, we employed an extractive text summarization approach~\cite{luhn1958,edmundson1969new,mihalcea2004textrank,erkan2004lexrank}. 
Each feedback text, either from the LLM or a human, was processed by GPT-4 to extract a list of the points of comments raised in the text (see prompt in \textbf{Supp. Fig.~\ref{fig:extract-comment-prompt}}). The output was structured in a JSON (JavaScript Object Notation) format. Within this format, each JSON key assigns an ID to a specific point, while the corresponding value details the content of the point (\textbf{Supp. Fig.~\ref{fig:extract-comment-prompt}}).
We focused on criticisms in the feedback, as they provide direct feedback to help authors improve their papers~\cite{goodman1994manuscript}.
The second stage focused on semantic text matching~\cite{deerwester1990indexing,socher2011dynamic,bowman2015large}. 
Here, we input both the JSON-formatted feedback from the LLM and the human into GPT-4. The LLM then generated another JSON output where each key identified a pair of matching point IDs and the associated value provided the explanation for the match. 
Given that our preliminary experiments showed GPT-4's matching to be lenient, we introduced a similarity rating mechanism. 
In addition to identifying corresponding pairs of matched comments, GPT-4 was also tasked with self-assessing match similarities on a scale from 5 to 10 (\textbf{Supp. Fig.~\ref{fig:match-comment-prompt}}). We observed that matches graded as ``5. Somewhat Related'' or ``6. Moderately Related'' introduced variability that did not always align with human evaluations. Therefore, we only retained matches ranked ``7. Strongly Related'' or above for subsequent analyses.

We validated our retrospective comment matching pipeline using human verification. 
In the extractive text summarization stage, we randomly selected 639 pieces of scientific feedback, including 150 from the LLM and 489 from human contributors. Two co-authors assessed each feedback and its corresponding list of extracted comments, identifying true positives (correctly extracted comments), false negatives (missed relevant comments), and false positives (incorrectly extracted or split comments). This process resulted in an F1 score of 0.968, with a precision of 0.977 and a recall of 0.960 (\textbf{Supp. Table~\ref{tab:combined_verification}$a$}), demonstrating the accuracy of the extractive summarization stage. 
For the semantic text matching stage, we sampled 760 pairs of scientific feedbacks: 332 comparing GPT to Human feedback and 428 comparing Human feedbacks. Each feedback pair was processed to enumerate all potential pairings of their extracted comment lists, resulting in 12,035 comment pairs. Three co-authors independently determined whether the comment pairs matched, without referencing the pipeline's predictions. Comparing these annotations with pipeline outputs yielded an F1 score of 0.824, a recall of 0.878, and a precision of 0.777 (\textbf{Supp. Table~\ref{tab:combined_verification}$b$}).
To assess inter-annotator agreement, we collected three annotations for 800 randomly selected comment pairs. Given the prevalence of non-matches, we employed stratified sampling, drawing 400 pairs identified as matches by the pipeline and 400 as non-matches. We then calculated pairwise agreement between annotations and the F1 score for each annotation against the majority consensus. The data showed 89.8\% pairwise agreement and an F1 score of 88.7\%, indicating the reliability of the semantic text matching stage.

\subsection*{Evaluating Specificity of LLM Feedback through Review Shuffling}

To evaluate the specificity of the feedback generated by the LLM, we compared the overlap between human-authored feedback and shuffled LLM feedback. 
For papers published in the \emph{Nature} journal family, the LLM-generated feedback for a given paper was randomly paired with human feedback for a different paper from the same journal and \emph{Nature} root category. These categories included physical sciences, earth and environmental sciences, biological sciences, health sciences, and scientific community and society. If a paper was classified under multiple categories, the shuffle algorithm paired it with another paper that spanned the same categories. For the \emph{ICLR} dataset, we compared human feedback for a paper with LLM feedback for a different paper, randomly selected from the same conference year, either \emph{ICLR} 2022 or \emph{ICLR} 2023. 
This shuffling procedure was designed to test the null hypothesis: 
if LLM mostly produces generic feedback applicable to many papers, 
then there would be little drop in the pairwise overlap between LLM feedback and the comments from each individual reviewer after the shuffling.

\subsection*{Overlap Metrics for Retrospective Evaluations and Control}

In the retrospective evaluation, we assessed the pairwise overlap of both GPT-4 vs. Human and Human vs. Human in terms of hit rate (\textbf{Fig.~\ref{fig:Main-AutoEval}}). The hit rate, defined as the proportion of comments in set \(A\) that match those in set \(B\), was calculated as follows: 
    \[
    \text{{Hit Rate}} = \frac{{|A \cap B|}}{{|A|}}
    \] 
To facilitate a direct comparison between the hit rates of GPT-4 vs. Human and Human vs. Human, we controlled for the number of comments when measuring the hit rate for Human vs. Human. Specifically, we considered only the first $N$ comments made by the first human (i.e., the human comments used as set \(A\)) for matching, where $N$ is the number of comments made by GPT-4 for the same paper. 
The results, with and without this control, were largely similar across both the \emph{ICLR} dataset for different decision outcomes (\textbf{Supp. Fig.~\ref{fig:supplementary-ICLR-single-stratify},\ref{fig:supplementary-scatter}}) and the \emph{Nature} family journals dataset across different journals (\textbf{Supp. Fig.~\ref{fig:supplementary-Nature-main-single-stratify},\ref{fig:supplementary-Nature-comm-single-stratify},\ref{fig:supplementary-scatter}}).
To examine the robustness of the results across different set overlap metrics, we also evaluated three additional metrics: the Szymkiewicz–Simpson overlap coefficient, the Jaccard index, and the Sørensen–Dice coefficient. These were calculated as follows: 
\begin{align*}
    \text{Szymkiewicz–Simpson Overlap Coefficient} &= \frac{|A \cap B|}{\min(|A|, |B|)} \\
    \text{Jaccard Index} &= \frac{|A \cap B|}{|A \cup B|} \\
    \text{Sørensen–Dice Coefficient} &= \frac{2|A \cap B|}{|A| + |B|}
\end{align*}
Results on these additional metrics suggest that our findings are robust on different set overlap metrics: the overlap GPT-4 vs. Human appears comparable to those of Human vs. Human both with and without control for the number of comments (\textbf{Supp. Fig.~\ref{fig:supplementary-metrics-vertical}}).

\subsection*{Characterizing the comment aspects in human and LLM feedback}

We curated an annotation schema of 11 key aspects to identify and measure the prevalence of these aspects in human and LLM feedback. This schema was developed with a focus on the \emph{ICLR} dataset, due to its specialized emphasis on Machine Learning. Each aspect was defined by its underlying emphasis, such as novelty, research implications, suggestions for additional experiments, and more. The selection of these 11 key aspects was based on a combination of the common schemes identified in the literature within the machine learning domain~\cite{values-in-ML-research,smith2022real,koch2021reduced,scheuerman2021datasets}, comments from machine learning researchers, and initial exploration by the annotators. 
From the \emph{ICLR} dataset, a random sample of 500 papers was selected to ensure a broad yet manageable representation. Using our extractive text summarization pipeline, we extracted lists of comments from both the LLM and human feedback for each paper. Each comment was then annotated according to our predefined schema, identifying any of the 11 aspects it represented (\textbf{Supp. Table~\ref{tab:values_ICLR_1},\ref{tab:values_ICLR_2},\ref{tab:values_ICLR_3}}). To ensure annotation reliability, two researchers with a background in machine learning performed the annotations.

\subsection*{Prospective User Study and Survey}

We conduct a prospective user study to further validate the effectiveness of leveraging LLMs to generate scientific feedback. To facilitate our user study, we launched an online Gradio demo~\cite{gradio} of the aforementioned generation pipeline, accessible at a public URL (\textbf{Supp. Fig.~\ref{fig:gradio}}). Users are prompted to upload a research paper in its original PDF format, after which the system delivers the review to user's email. 
We ask users to only upload papers published after 9/2021 to ensure the papers are never seen by GPT-4 during training (the cutdown date of GPT-4 training corpora is 9/2021). We have also incorporated an ethics statement to discourage the direct use of LLM content for any review-related tasks.
After the review is generated and sent to users, users are asked to fill a 6-page survey (Figure~\ref{fig:user_study}), which includes 1) author background information, 2) review situation in author's area, 3) general impression of LLM review, 4) detailed evaluation of LLM review, 5) comparison with human review, and 6) additional questions and feedback, which systematically investigates human evaluations of different aspects of LLM reviews. The survey takes around 15-20 minutes and users will be compensated with \$20. 
We recruit the participants through 1) relevant institute mailing lists, and 2) reaching out to all authors who have published at least one preprint on arXiv in the field of computer science and computational biology during January to March, 2023, provided their email contact information is available in the first three pages of the PDF. The study has been approved by Stanford University's Institutional Review Board.

\section*{Acknowledgements} 
We thank S. Eyuboglu, D. Jurafsky, and M. Bernstein for their guidance and helpful discussions.
J.Z. is supported by the National Science Foundation (CCF 1763191 and CAREER 1942926), the US National Institutes of Health (P30AG059307 and U01MH098953) and grants from the Silicon Valley Foundation and the Chan-Zuckerberg Initiative. H.C. is supported by the Stanford Interdisciplinary Graduate Fellowship.

\section*{Code Availability}
The codes can be accessed at \url{https://github.com/Weixin-Liang/LLM-scientific-feedback}.

\bibliography{ref}

\clearpage 
\newpage

\clearpage

\newcounter{mysuppfigure}
\newenvironment{mysuppfigure}[1][]{%
  \addtocounter{mysuppfigure}{1}%
  \renewcommand{\thesuppfigure}{S\arabic{mysuppfigure}}%
  \begin{suppfigure}[#1]%
}{%
  \end{suppfigure}%
}

\newcounter{mysupptable}
\newenvironment{mysupptable}[1][]{%
  \addtocounter{mysupptable}{1}%
  \renewcommand{\thesupptable}{S\arabic{mysupptable}}%
  \begin{supptable}[#1]%
}{%
  \end{supptable}%
}

\clearpage
\newpage 

\appendix
\section*{Supplementary Information}

\subsection*{Additional Related Work}

The use of AI tools in aiding the scientific publication process has garnered considerable attention. Algorithms have been developed to summarize paper contents~\cite{collins2017supervised}, detect inaccurately reported p-values~\cite{nuijten2016prevalence}, rectify citation errors~\cite{aclpubcheck}, and identify fairness disparities~\cite{zhang2022investigating}. 
Recent advances in LLMs like ChatGPT and GPT-4 have intensified interest in leveraging these technologies for scientific feedback.
There are some exploratory and unpublished studies. Hosseini et al. conducted a small-scale qualitative investigation to gauge ChatGPT's effectiveness in the peer review process~\cite{hosseini2023fighting}. Similarly, Robertson et al. involved 10 participants to assess GPT-4's benefits in aiding peer review~\cite{robertson2023gpt4}. Liu et al. demonstrated that GPT-4 could identify paper errors, verify checklists, and compare paper quality through analysis of 10-20 computer science papers~\cite{liu2023reviewergpt}. Verharen et al. utilized ChatGPT to analyze 200 published neuroscience papers and uncovered gender disparities in scientific peer review~\cite{verharen2023chatgpt}.

Our study differs from the existing literature in two key aspects.  
First, we provide a large-scale empirical analysis, in contrast to the small-scale exploratory analysis conducted in previous efforts. 
Our dataset includes 3,096 papers from the \emph{Nature} family of journals and 1,709 papers from the \emph{ICLR} conference, spanning a wide range of scientific disciplines. 
We also incorporate 308 human responses from 110 US institutions obtained through a prospective user study. 
Secondly, while most previous works only present qualitative results, our work provides a systematic quantitative assessment. This involves both overlap analyses and human evaluation metrics, providing a comprehensive analysis of the promise and limitations of LLMs in providing scientific feedback.

\clearpage 
\newpage

\begin{supptable}
\centering
\caption{Summary of papers and their associated reviews sampled from 15 \emph{Nature} family journals.}
\label{tab:nature_data}
\begin{tabular}{lcc}
\toprule
\multirow{2}{*}{\textbf{Journal}} & \multicolumn{2}{c}{\textbf{Count}} \\
\cmidrule(lr){2-3}
 & \textbf{Papers} & \textbf{Reviews} \\
\midrule
\emph{Nature} & 773 & 2324 \\
\emph{Nature Communications} & 810 & 2250 \\
\emph{Communications Earth \& Environment} & 299 & 807 \\
\emph{Communications Biology} & 200 & 526 \\
\emph{Communications Physics} & 174 & 464 \\
\emph{Communications Medicine} & 123 & 343 \\
\emph{Nature Ecology \& Evolution} & 113 & 332 \\
\emph{Nature Structural \& Molecular Biology} & 110 & 290 \\
\emph{Communications Chemistry} & 101 & 279 \\
\emph{Nature Cell Biology} & 78 & 233 \\
\emph{Nature Human Behaviour} & 72 & 211 \\
\emph{Communications Materials} & 67 & 181 \\
\emph{Nature Immunology} & 62 & 165 \\
\emph{Nature Microbiology} & 57 & 174 \\
\emph{Nature Biomedical Engineering} & 57 & 166 \\
\midrule
Total & 3096 & 8745 \\
\bottomrule
\end{tabular}
\end{supptable}

\clearpage 
\newpage

\begin{supptable}
\centering
\caption{Summary of \emph{ICLR} papers and their associated reviews sampled from the years 2022 and 2023, grouped by decision.}
\label{tab:iclr_data}
\resizebox{0.95\textwidth}{!}{%
\begin{tabular}{lcccc}
\toprule
\multirow{2}{*}{\textbf{Decision Category}} & \multicolumn{2}{c}{\textbf{\emph{ICLR} 2022}} & \multicolumn{2}{c}{\textbf{\emph{ICLR} 2023}} \\
\cmidrule(lr){2-3} \cmidrule(lr){4-5}
 & \# of Papers & \# of Reviews & \# of Papers & \# of Reviews \\
\midrule
Accept (Oral) - notable-top-5\% & 55 & 200 & 90 & 317 \\
Accept (Spotlight) - notable-top-25\% & 173 & 664 & 200 & 758 \\
Accept (Poster) & 197 & 752 & 200 & 760 \\
Reject after author rebuttal & 213 & 842 & 212 & 799 \\
Withdrawn after reviews & 182 & 710 & 187 & 703 \\
\midrule
Total & 820 & 3168 & 889 & 3337 \\
\bottomrule
\end{tabular}
}
\end{supptable}

\clearpage 
\newpage

\begin{supptable}
\centering
\caption{
\textbf{Results of human verification on the retrospective comment extraction and matching pipeline.} 
\textbf{a}, 
Human verification for the extractive summarization stage. 
We randomly sampled 639 scientific feedbacks. 
Human annotators reported the numbers of true positives (correctly extracted comments), false negatives (overlooked relevant comments), and false positives (incorrectly extracted or split comments).
Results suggest high accuracy of the extractive summarization stage.
\textbf{b}, 
Human verification for the semantic matching stage.
We randomly sampled 12,035 pairs of extracted comments. Human annotators annotated each pair of extracted comments, determining whether the two comments are semantically matched in content. 
Results suggest high accuracy of the semantic text matching stage.
}
\label{tab:combined_verification}
\begin{minipage}{0.5\textwidth}
\centering
\subcaption{Extractive Summarization}
\begin{threeparttable}
\begin{tabular}{@{}lc@{}}
\toprule
\textbf{Extracted Comments} & \textbf{Count} \\
\midrule
TP (True Positives) & \cellcolor{lightgray}2634 \\
FN (False Negatives) & 110 \\
FP (False Positives) & 63 \\
\midrule
\textbf{Precision} & 0.977 \\
\textbf{Recall} & 0.960 \\
\textbf{F1 Score} & 0.968 \\
\bottomrule
\end{tabular}
\end{threeparttable}
\end{minipage}%
\begin{minipage}{0.5\textwidth}
\centering
\subcaption{Semantic Matching}
\begin{threeparttable}
\begin{tabular}{@{}lcc@{}}
\toprule
 & \multicolumn{2}{c}{\textbf{Predicted Matching}} \\
\cmidrule(lr){2-3}
\textbf{Human Matching} & Matched & Not Matched \\
\midrule
Matched & \cellcolor{lightgray}685 & 95 \\
Not Matched & 197 & \cellcolor{lightgray}11058 \\
\midrule
\textbf{Precision} & 0.777 & \\
\textbf{Recall} & 0.878 & \\
\textbf{F1 Score} & 0.824 & \\
\bottomrule
\end{tabular}
\end{threeparttable}
\end{minipage}
\end{supptable}

\clearpage 
\newpage 

\begin{supptable}[ht]
\centering
\caption{Mean token lengths of papers and human reviews in the two datasets.}
\label{tab:paper_review_length}
\begin{tabular}{lcc}
\toprule
\textbf{Dataset} & \textbf{Paper (Mean Token Length)} & \textbf{Human Review (Mean Token Length)} \\
\midrule
\emph{ICLR} & 5,841.46 & 671.53 \\
\emph{Nature} Family Journals & 12,444.06 & 1,337.93 \\
\bottomrule
\end{tabular}
\end{supptable}

\clearpage 
\newpage 

\begin{supptable}[ht]
\centering
\caption{Example comments extracted from LLM and human feedback on \emph{ICLR} by human coding}
\label{tab:values_ICLR_1}
\begin{tabulary}{0.95\textwidth}{lcL@{}}
\toprule
\textbf{Human Coding} & \textbf{Source} & \textbf{Comment} \\
\midrule
Clarity and Presentation & Human & The writing is hard to follow. Since this paper introduces multiple new concepts, it was hard for me to understand... \\
\midrule
Clarity and Presentation & GPT-4 & The paper is highly technical and may be difficult to understand for readers who are not familiar with the field. The authors could provide a more detailed explanation of the IB principle... \\
\midrule
Comparison to Previous Studies & Human & 
The comparisons are flawed. In particular the 'label consistency' and 'class-center consistency' losses are disjoint with the GNN methodology, and a fairer comparison would be with GNNs that also use these two losses... \\
\midrule
Comparison to Previous Studies & GPT-4 & The paper lacks a thorough comparison with existing methods. While the authors have compared their approach with a few baselines, a more comprehensive comparison with... \\
\midrule
Theoretical Soundness & Human & IMHO, the theoretical proof is relatively trivial. The final conclusion is if the similarity is proper, the predicted action is accurate. Since the model is actually learning the proper similarity, this is equivalent to saying if the model $h$ is well trained, the output is accurate. This is obviously true. \\
\midrule
Theoretical Soundness & GPT-4 & The authors should provide more details on the theoretical analysis of the connections between message passing and consistency constraints to make it more understandable for readers... \\
\midrule
Novelty & Human & A major concern of the paper is about the model's novelty. The reviewer has doubts on the argument that a new combination of existing techniques (BN, LSTM, S\&E, skip connection) for the task of video prediction is significant enough to publish in \emph{ICLR}... \\
\midrule
Novelty & GPT-4 & The paper could be more explicit in explaining the novelty of the proposed method compared to existing techniques. It is not entirely clear how the proposed method differs from other methods that use projection features...  \\
\midrule
Reproducibility & Human & As for reproducibility, although the authors provide a reproducibility statement, it is still better to provide codes for other readers to implement the experiments... \\
\midrule
Reproducibility & GPT-4 & Reproducibility: The authors should provide more details about the experimental setup to ensure reproducibility... \\
\bottomrule
\end{tabulary}
\end{supptable}

\clearpage 
\newpage 

\begin{supptable}[ht]
\centering
\caption{Example comments extracted from LLM and human feedback on \emph{ICLR} by human coding}
\label{tab:values_ICLR_2}
\begin{tabulary}{0.95\textwidth}{lcL@{}}
\toprule
\textbf{Human Coding} & \textbf{Source} & \textbf{Comment} \\
\midrule
Add ablations experiments & Human & It's best to make clearer how much of the observed performance gains in experiments is from the proposed learning confidence-conditioned values, vs from some of the empirical/engineering decisions in Sec 5 practical algorithms (IQN, approx inverse visitation). Perhaps if there's a simpler (maybe tabular?) problem that doesn't have these deep RL complexities that would strengthen the results and help the reader understand the contribution better - and help to show what the learned Q-values are actually doing... \\
\midrule
Add ablations experiments & GPT-4 & The paper does not provide a clear explanation of how the proposed auxiliary tasks contribute to the improved performance of the model. The authors should provide a more detailed analysis of the impact of these tasks on the model's performance. It would also be helpful to see a comparison of the model's performance with and without these tasks... \\
\midrule
Implications of the Research & Human & One aspect I would have liked to see more of is how the findings of this submission influence future work on Gaussian processes and other Bayesian models? Are there any specific takeaways which might guide the design of models yielding improved estimates with better uncertainty calibration? \\
\midrule
Implications of the Research & GPT-4 & The paper could provide more information on the potential real-world implications of membership leakage in PLMs. For example: The authors could discuss the potential impact on users' privacy and data security. The authors could explore the potential legal and ethical implications of such privacy leakage... \\
\midrule
Ethical Aspects & Human & My main concern: I did not find IRB approval information on the human experiment. If there is, it should be mentioned, if not the authors should explain why it is not necessary in this case (and should be validated with the conference chairs). Also, the details of the experiment and instructions to the demonstrators should accompany the paper. \\
\midrule
Ethical Aspects & GPT-4 & The paper does not discuss the ethical implications of their research. While the authors' intention is to improve the security of federated learning systems, their research could potentially be misused by malicious actors... \\
\bottomrule
\end{tabulary}
\end{supptable}

\clearpage 
\newpage 

\begin{supptable}[ht]
\centering
\caption{Example comments extracted from LLM and human feedback on \emph{ICLR} by human coding}
\label{tab:values_ICLR_3}
\begin{tabulary}{0.95\textwidth}{lcL@{}}
\toprule
\textbf{Human Coding} & \textbf{Source} & \textbf{Comment} \\
\midrule
Add ablations experiments & Human & It's best to make clearer how much of the observed performance gains in experiments is from the proposed learning confidence-conditioned values, vs from some of the empirical/engineering decisions in Sec 5 practical algorithms (IQN, approx inverse visitation). Perhaps if there's a simpler (maybe tabular?) problem that doesn't have these deep RL complexities that would strengthen the results and help the reader understand the contribution better - and help to show what the learned Q-values are actually doing... \\
\midrule
Add ablations experiments & GPT-4 & The paper does not provide a clear explanation of how the proposed auxiliary tasks contribute to the improved performance of the model. The authors should provide a more detailed analysis of the impact of these tasks on the model's performance. It would also be helpful to see a comparison of the model's performance with and without these tasks... \\
\midrule
Implications of the Research & Human & One aspect I would have liked to see more of is how the findings of this submission influence future work on Gaussian processes and other Bayesian models? Are there any specific takeaways which might guide the design of models yielding improved estimates with better uncertainty calibration? \\
\midrule
Implications of the Research & GPT-4 & The paper could provide more information on the potential real-world implications of membership leakage in PLMs. For example: The authors could discuss the potential impact on users' privacy and data security. The authors could explore the potential legal and ethical implications of such privacy leakage... \\
\midrule
Ethical Aspects & Human & My main concern: I did not find IRB approval information on the human experiment. If there is, it should be mentioned, if not the authors should explain why it is not necessary in this case (and should be validated with the conference chairs). Also, the details of the experiment and instructions to the demonstrators should accompany the paper. \\
\midrule
Ethical Aspects & GPT-4 & The paper does not discuss the ethical implications of their research. While the authors' intention is to improve the security of federated learning systems, their research could potentially be misused by malicious actors... \\
\bottomrule
\end{tabulary}
\end{supptable}

\clearpage 
\newpage

\begin{suppfigure*}[htb]  
\centering
\includegraphics[width=0.75\textwidth]{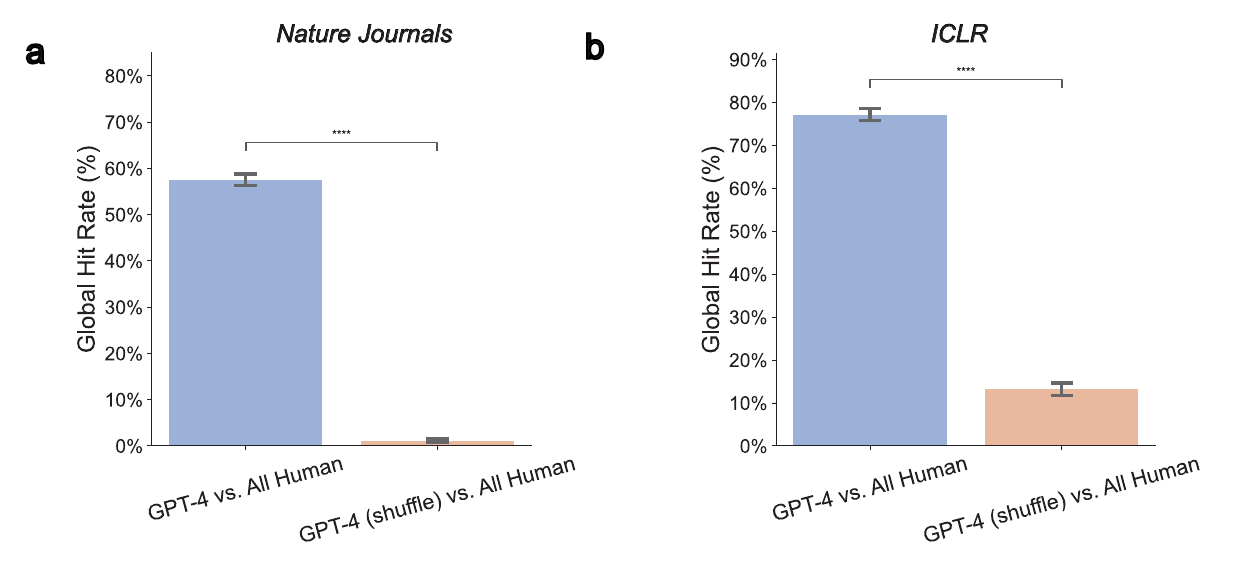}
\caption{
\textbf{Fraction of GPT-4 comments that overlap with comments raised by at least one human reviewer.}
\textbf{a}, In the \emph{Nature} family journal data, 57.55\% of the comments made by GPT-4 overlap with comments made by at least one human reviewer, suggesting a significant overlap between LLM feedback and human feedback. 
“GPT-4 (shuffle)” refers to feedback from GPT-4 for a randomly selected manuscript from the same journal and category. 
As a null model, if LLM mostly produces generic feedback applicable to many papers, 
then there would be little drop in the global hit rate between LLM feedback and the comments from each individual reviewer after the shuffling.
However, the global hit rate decreases markedly from 57.55\% to 1.13\% after shuffling, indicating that the LLM feedback is paper-specific. 
\textbf{b}, The results are similar with \emph{ICLR} data.
77.18\% of the comments made by GPT-4 overlap with comments made by at least one human reviewer. 
The shuffling result also suggests that LLM feedback is paper-specific. 
Error bars represent 95\% confidence intervals. 
*P < 0.05, **P < 0.01, ***P < 0.001, and ****P < 0.0001.
}
\label{fig:supplementary-global-hit-rates}
\end{suppfigure*}

\clearpage 
\newpage 

\begin{suppfigure*}[htb]  
\centering
\includegraphics[width=0.6\textwidth]{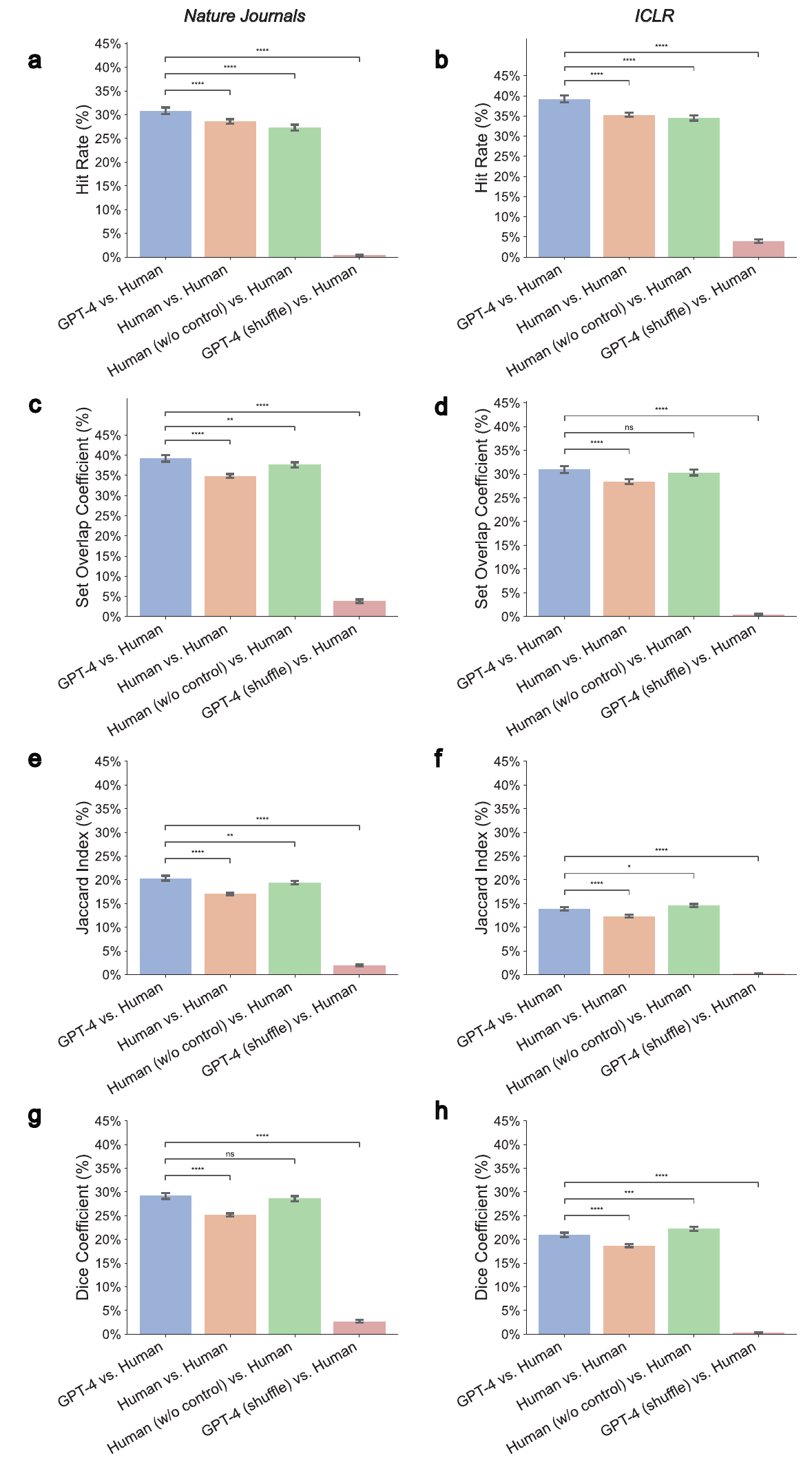}
\caption{
\textbf{Retrospective evaluation using alternative set overlap metrics for robustness check.}
\textbf{(a, b)} Hit rate.
\textbf{(c, d)} Szymkiewicz–Simpson overlap coefficient.
\textbf{(e, f)} Jaccard index.
\textbf{(g, h)} Sørensen–Dice coefficient.
Additional metrics results suggest the overlap between GPT-4 and Human is comparable to Human vs. Human overlap, suggesting the robustness of our findings across different set overlap metrics.  
Error bars represent 95\% confidence intervals. 
*P < 0.05, **P < 0.01, ***P < 0.001, and ****P < 0.0001.
}
\label{fig:supplementary-metrics-vertical}
\end{suppfigure*}

\clearpage 
\newpage 

\begin{suppfigure*}[t!]
    \centering
    \includegraphics[width=0.7\linewidth]{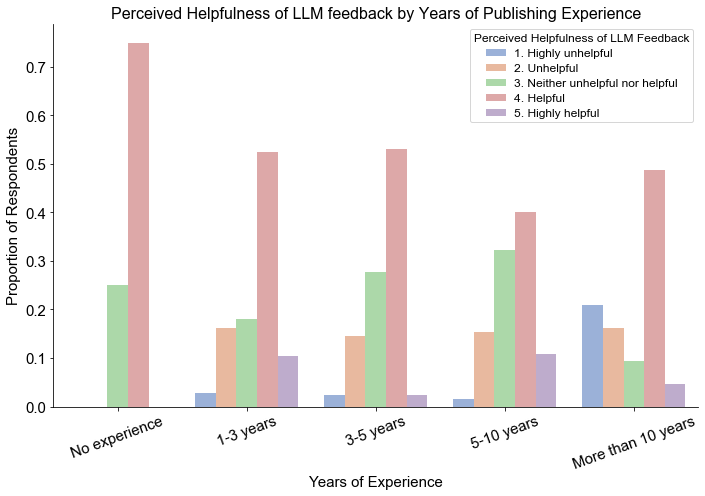}
    
    \caption{LLM-based scientific feedback is considered helpful among participants with varying publishing experience.}
    \label{fig:plots_helpfulness_years_pdf}
\end{suppfigure*}

\begin{suppfigure*}[t!]
    \centering
    \includegraphics[width=0.7\linewidth]{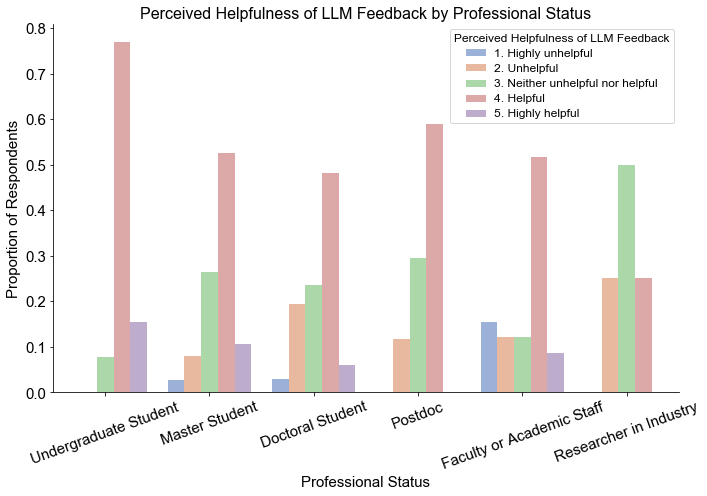}
    
    \caption{LLM-based scientific feedback is considered helpful among participants of different professional statuses.}
    \label{fig:plots_helpfulness_status_pdf}
\end{suppfigure*}

\clearpage 
\newpage 

\begin{suppfigure*}[t!]  
\centering
\includegraphics[width=\textwidth]{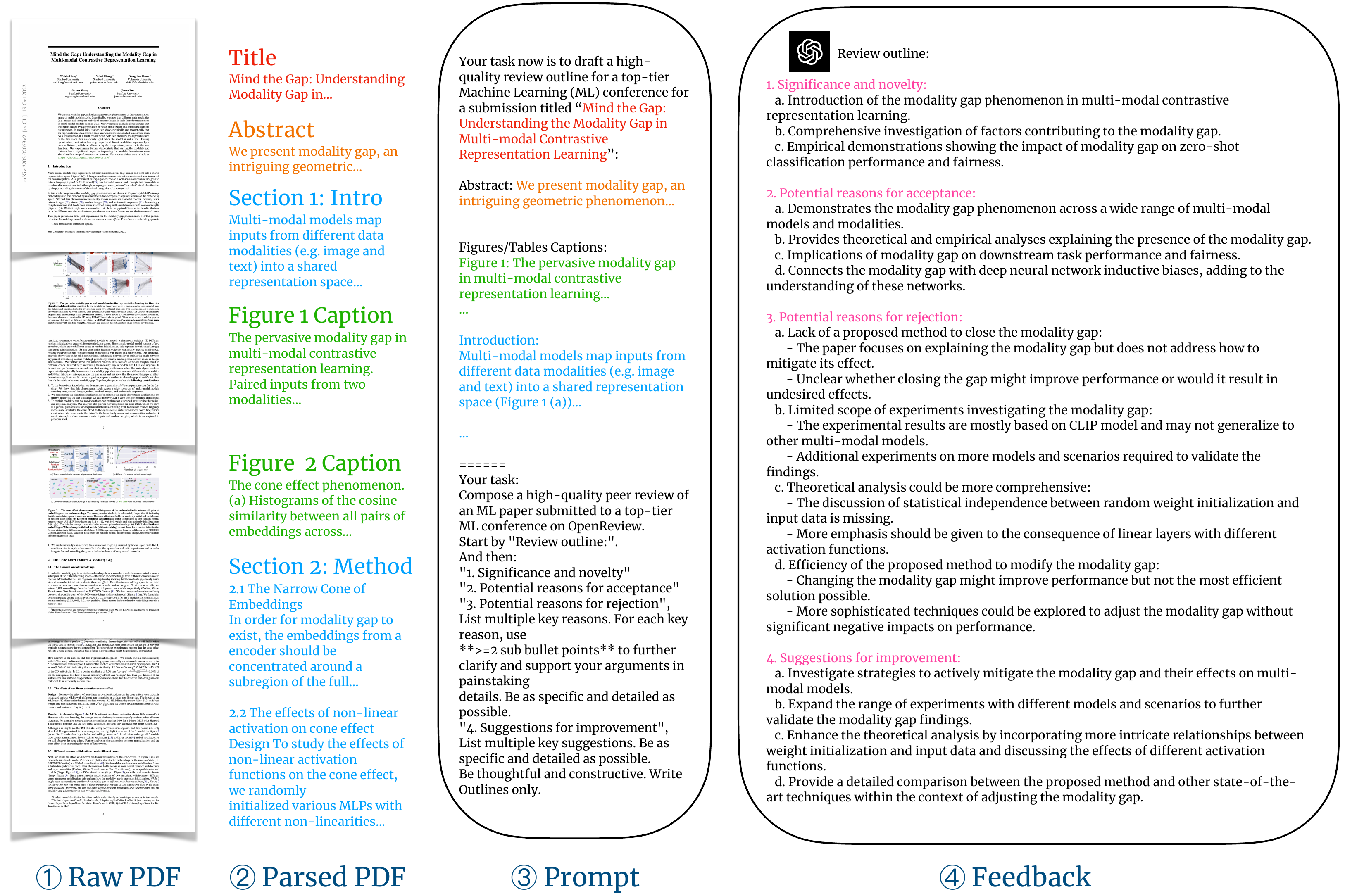}
\caption{
\textbf{Schematic of the LLM scientific feedback generation system.} Manuscript text, including figure captions, is extracted from the manuscript PDFs and integrated into a prompt for LLM GPT-4, which then generates feedback. 
The generated feedback provides structured comments in four sections: 
significance and novelty, 
potential reasons for acceptance, 
potential reasons for rejection, and 
suggestions for improvement.
In the example, GPT-4 raised a comment that the paper reported a modality gap phenomenon but did not propose methods to close the gap or demonstrate the benefits of doing so. 
}
\label{fig:supplementary-input_output}
\end{suppfigure*}

\clearpage 
\newpage

\begin{suppfigure*}[t!]  
\centering
\includegraphics[width=0.75\textwidth]{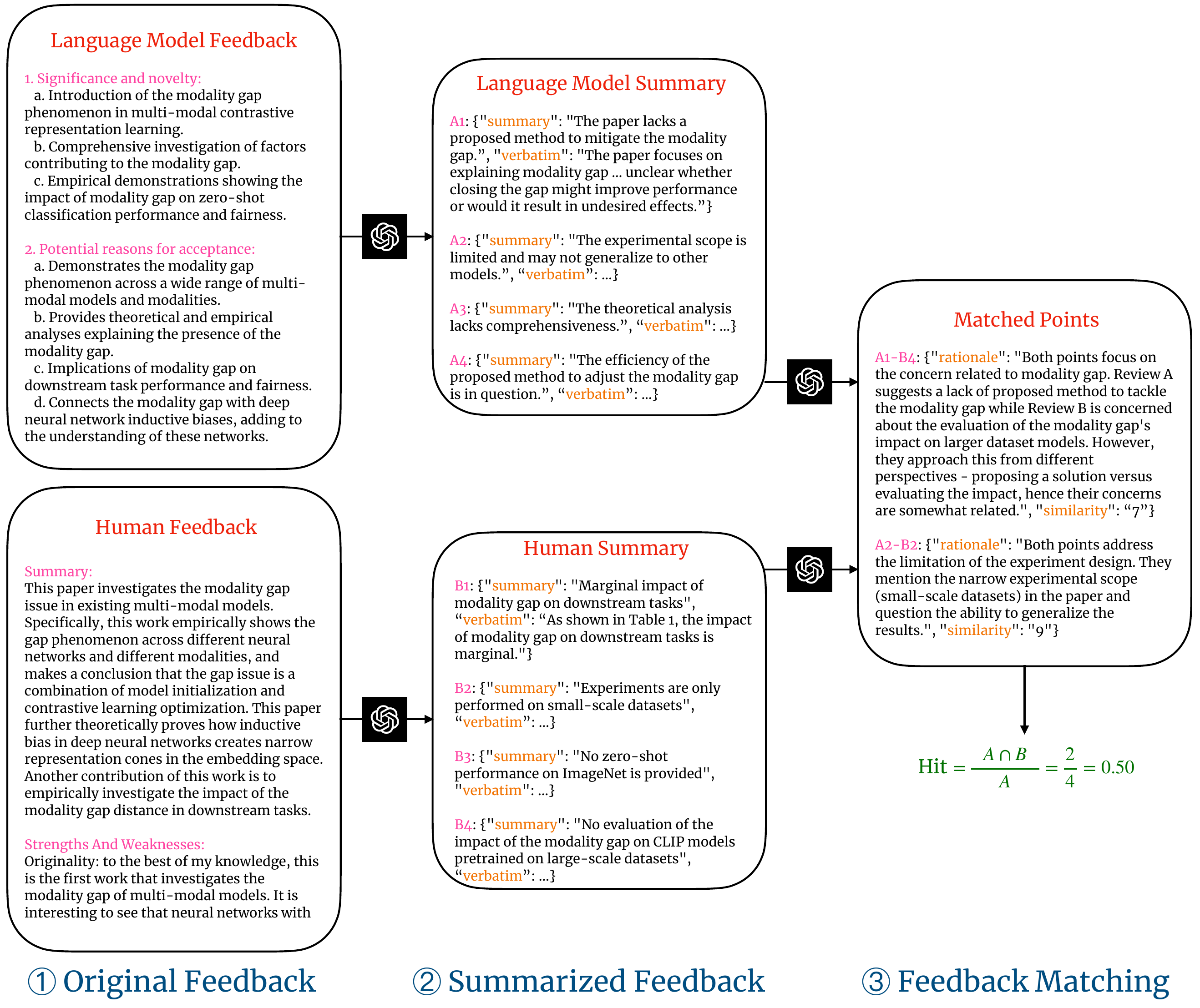}
\caption{
\textbf{Workflow of the retrospective comment matching pipeline for scientific feedback texts.}
\textbf{a}, This two-stage pipeline compares comments raised in LLM generated feedback with those from human reviewers.
\textbf{b}, Extraction: Utilizing LLM's capabilities for information extraction, key comments are extracted from both LLM generated and human-written reviews.
\textbf{c}, Matching: LLM is used for semantic similarity analysis, where comments from LLM and human feedback are matched. For each paired comment, a similarity rating and justifications are provided. A similarity threshold $\ge$ 7 is set to filter out weakly-matched comments. This threshold is chosen based on human validations of the matching stage. 
}
\label{fig:supplementary-retrospective_evaluation}
\end{suppfigure*}

\clearpage 
\newpage

\begin{suppfigure*}[htb]  
\centering
\includegraphics[width=1\textwidth]{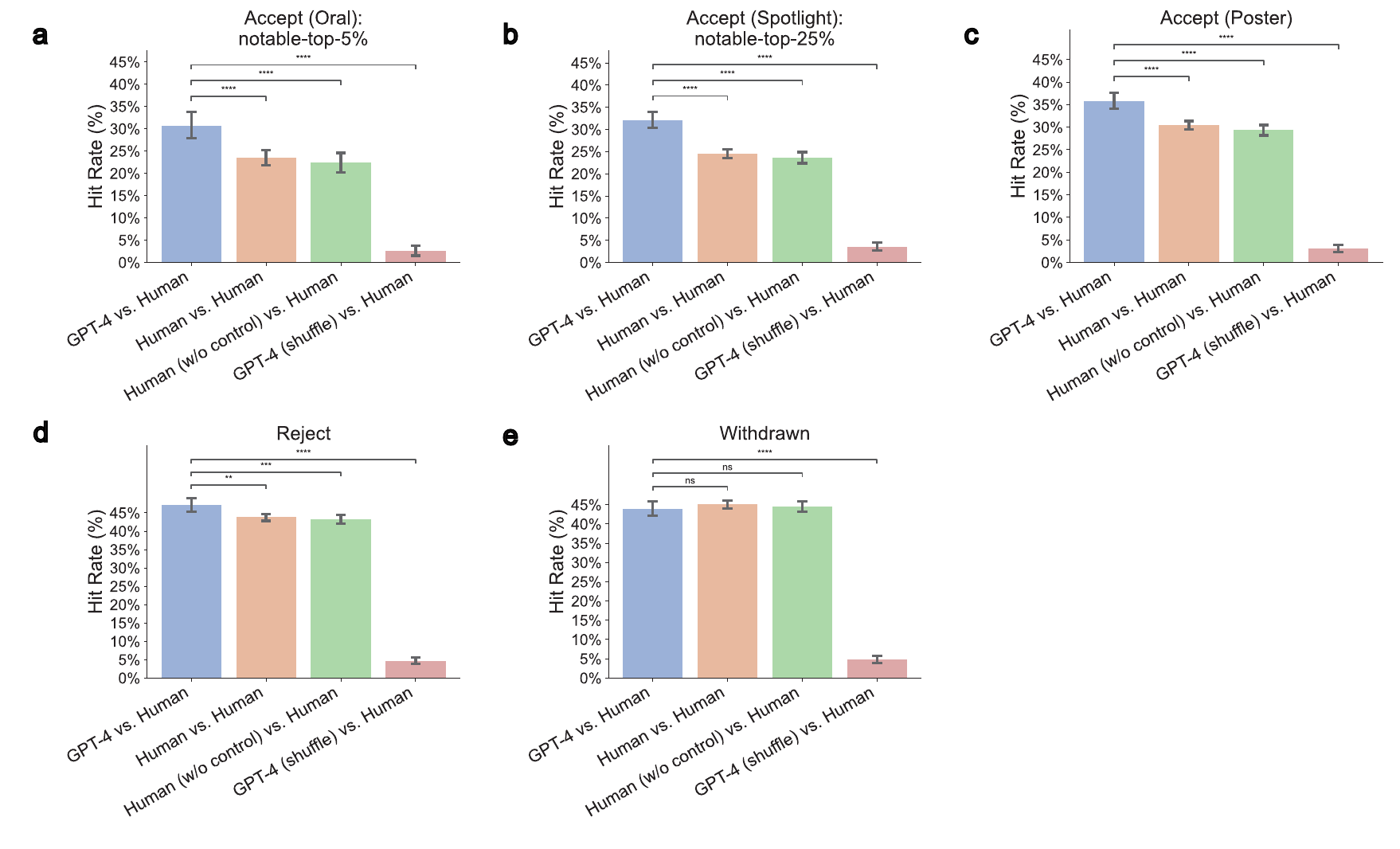}
\caption{
\textbf{Robustness check on controlling the number of comments for hit rate measurement in \emph{ICLR} data.}
Consider two sets of comments \(A\) and \(B\). The hit rate refers to the percentage of comments in set \(A\) that match those in set \(B\). 
To enable a more direct comparison between the hit rates of GPT-4 vs. Human and Human vs. Human, we have controlled for the number of comments when calculating the hit rate for Human vs. Human.  Specifically, we chose only the first $N$ comments made by the first reviewer (i.e., the comments used as set \(A\)) for matching, where $N$ corresponds to the number of comments made by GPT-4 for the same paper. 
We also included Human (w/o control) vs. Human, where we matched all human comments without limiting the number of comments. 
\textbf{a}, Hit rate comparison for \emph{ICLR} papers that were accepted with oral presentations (top 5\% accepted papers), 
\textbf{b}, papers accepted with spotlights (top 25\% of accepted papers),
\textbf{c}, papers accepted for poster presentations,
\textbf{d}, rejected papers,
\textbf{e}, papers withdrawn post-review. 
This figure indicates that results with and without the control for the number of comments are largely similar.
The overlap between LLM feedback and human feedback appears comparable to the overlap observed between two human reviewers.
Error bars represent 95\% confidence intervals. 
*P < 0.05, **P < 0.01, ***P < 0.001, and ****P < 0.0001.
}
\label{fig:supplementary-ICLR-single-stratify}
\end{suppfigure*}

\clearpage 
\newpage 

\begin{suppfigure*}[htb]  
\centering
\includegraphics[width=1\textwidth]{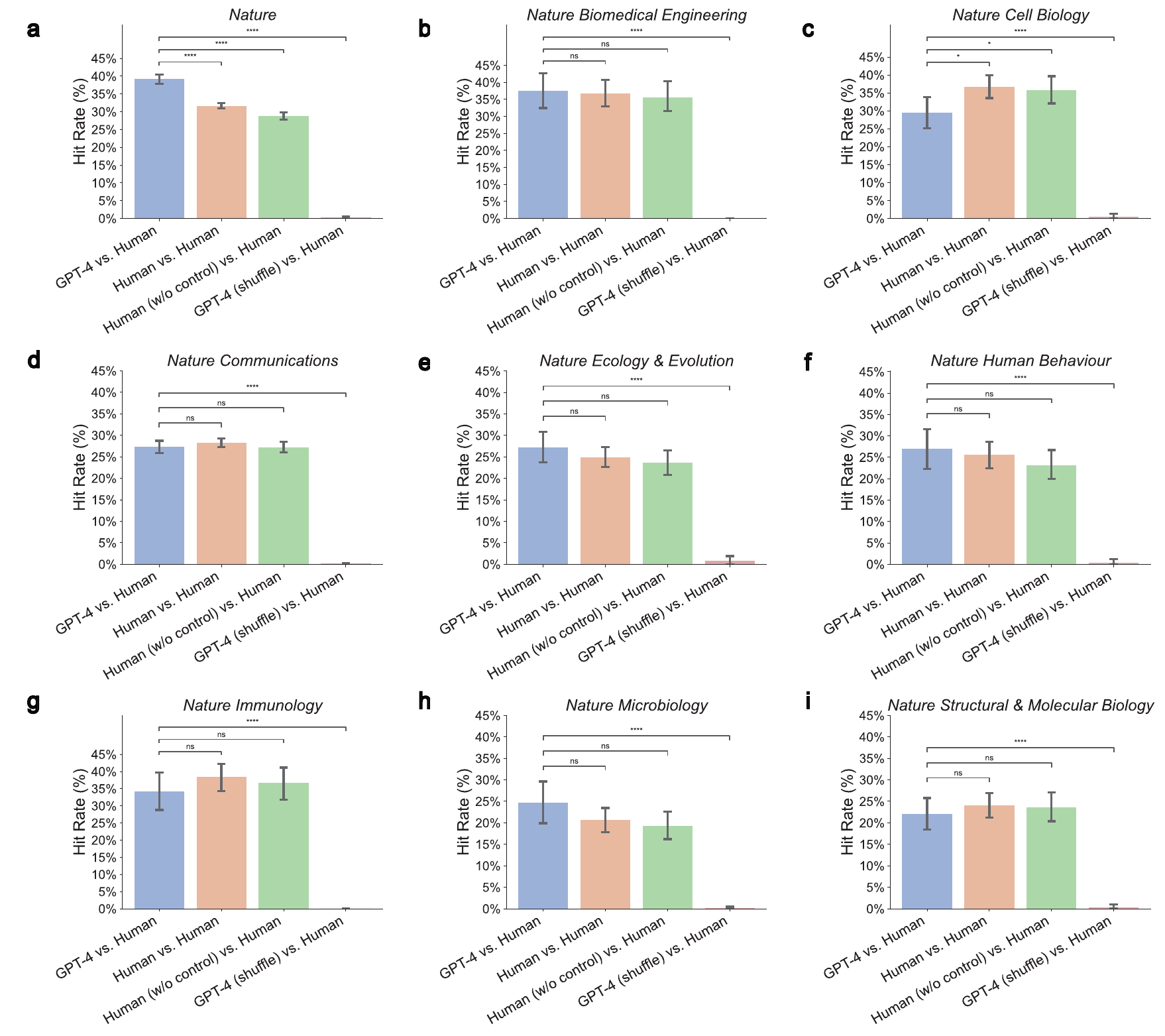}
\caption{
\textbf{Robustness check on controlling for the number of comments when measuring overlap in \emph{Nature} family journal data (1/2).}
Results are stratified by journals: 
(\textbf{a}) \emph{Nature}, 
(\textbf{b}) \emph{Nature Biomedical Engineering}, 
(\textbf{c}) \emph{Nature Cell Biology},
(\textbf{d}) \emph{Nature Ecology \& Evolution}, 
(\textbf{e}) \emph{Nature Human Behaviour}, 
(\textbf{f}) \emph{Nature Communications}, 
(\textbf{g}) \emph{Nature Immunology}, 
(\textbf{h}) \emph{Nature Microbiology}, 
and (\textbf{i}) \emph{Nature Structural \& Molecular Biology}. 
This figure indicates that results with and without the control for the number of comments are largely similar. The overlap between LLM feedback and human feedback seems comparable to the overlap observed between two human reviewers.
Results for additional \emph{Nature} family journals are shown in \textbf{Supp. Fig.~\ref{fig:supplementary-Nature-comm-single-stratify}}.  
Error bars represent 95\% confidence intervals. 
*P < 0.05, **P < 0.01, ***P < 0.001, and ****P < 0.0001.
}
\label{fig:supplementary-Nature-main-single-stratify}
\end{suppfigure*}

\clearpage 
\newpage 

\clearpage 
\newpage 

\begin{suppfigure*}[htb]  
\centering
\includegraphics[width=1\textwidth]{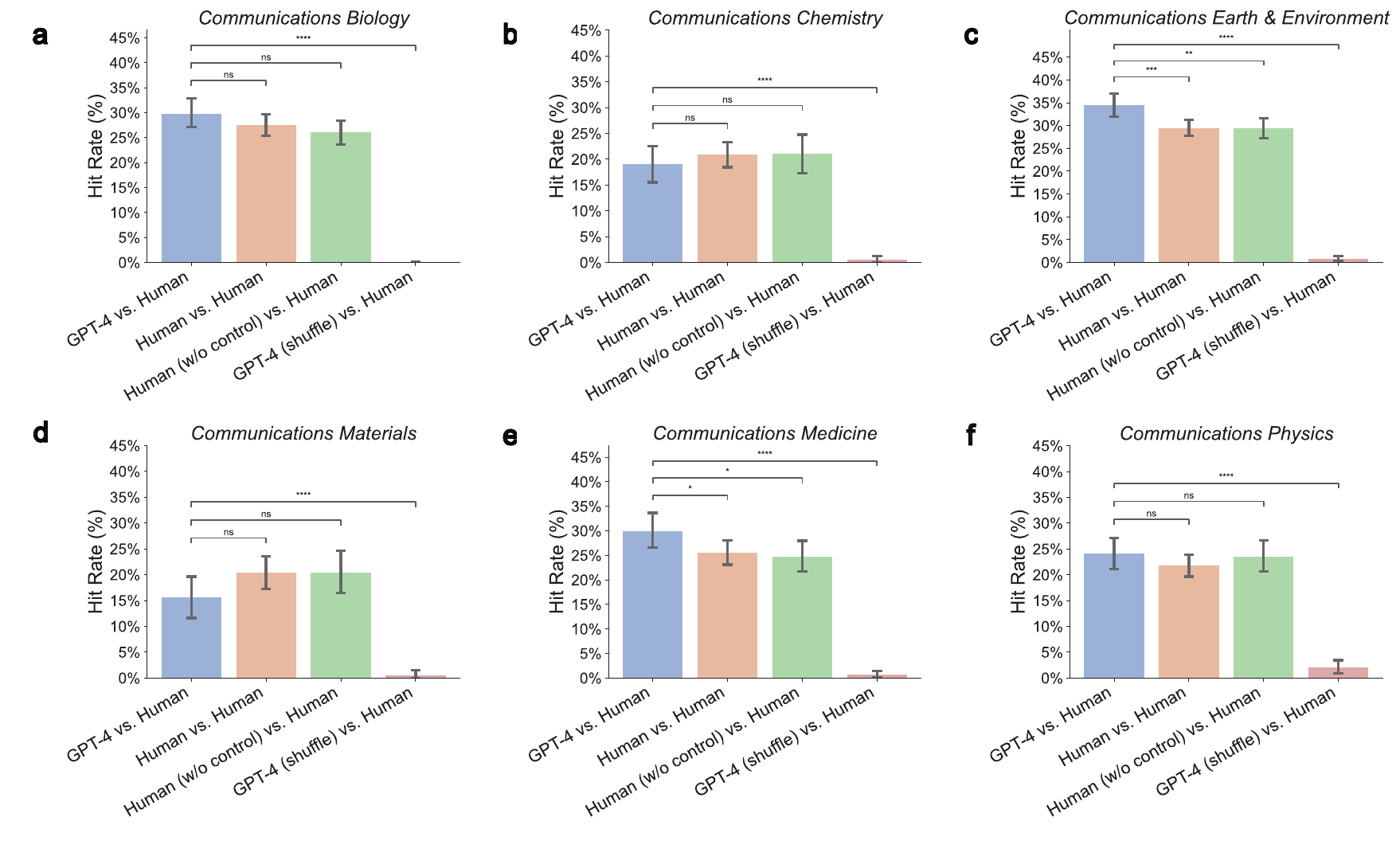}
\caption{
\textbf{Robustness check on controlling for the number of comments in measuring hit rates in \emph{Nature} family journal data (2/2).}
Results are stratified by journals (continuing \textbf{Supp. Fig.~\ref{fig:supplementary-Nature-main-single-stratify}}): 
(\textbf{a}) \emph{Communications Biology},  
(\textbf{b}) \emph{Communications Chemistry}, 
(\textbf{c}) \emph{Communications Earth \& Environment}, 
(\textbf{d}) \emph{Communications Materials}, 
(\textbf{e}) \emph{Communications Medicine}, 
and (\textbf{f}) \emph{Communications Physics}. 
This figure indicates that results with and without the control for the number of comments are largely similar. The overlap between LLM feedback and human feedback seems comparable to the overlap observed between two human reviewers.
Error bars represent 95\% confidence intervals. 
*P < 0.05, **P < 0.01, ***P < 0.001, and ****P < 0.0001.
}
\label{fig:supplementary-Nature-comm-single-stratify}
\end{suppfigure*}

\clearpage 
\newpage

\begin{suppfigure*}[htb]  
\centering
\includegraphics[width=1\textwidth]{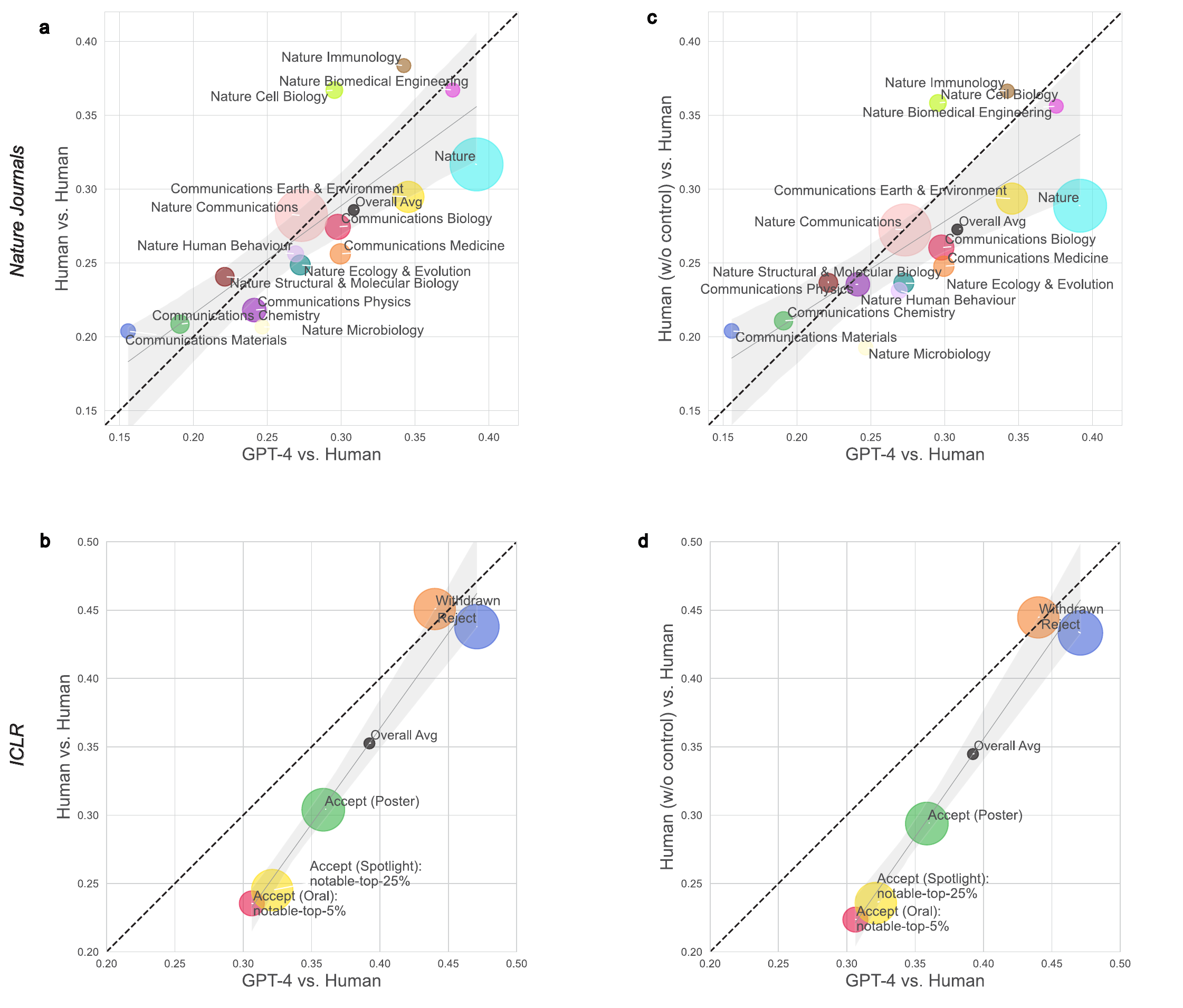}
\caption{
\textbf{Robustness check on controlling for the number of comments in the correlation of hit rates.}
\textbf{(a)} Hit rates across various \emph{Nature} family journals, controlling for the number of comments (\(r = 0.80\), \(P = 3.69 \times 10^{-4}\)). 
\textbf{(b)} Hit rates across \emph{ICLR} papers with different decision outcomes, controlling for the number of comments (\(r = 0.98\), \(P = 3.28 \times 10^{-3}\)). 
\textbf{(c)} Hit rates across various \emph{Nature} family journals without control for the number of comments (\(r = 0.75\), \(P = 1.37 \times 10^{-3}\)).
\textbf{(d)} Hit rates across \emph{ICLR} papers with different decision outcomes without control for the number of comments (\(r = 0.98\), \(P = 2.94 \times 10^{-3}\)). 
Human (w/o control) represents matching with all human comments without controlling for the number of comments. 
Circle size indicates sample size.
}
\label{fig:supplementary-scatter}
\end{suppfigure*}

\clearpage 
\newpage

\begin{suppfigure}
    \centering
    \includegraphics[width=\linewidth]{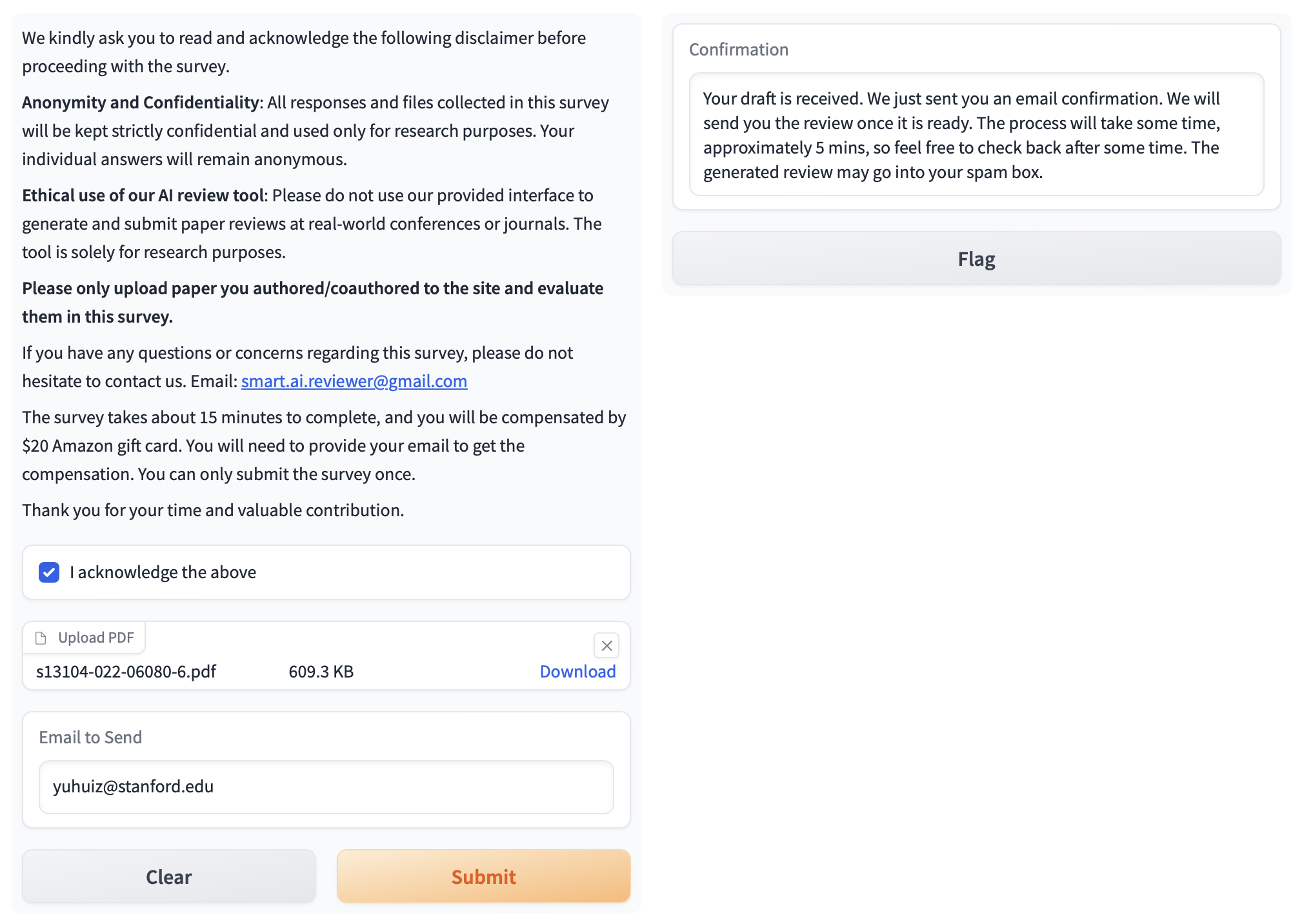}
\caption{Web interface for our prospective user study designed to characterize the capability of LLM in providing helpful scientific feedback. Users are instructed to upload a research paper they authored in PDF format, following which LLM feedback is sent to their email. Users were guided to upload only papers published after GPT-4's last training cut-off in September 2021. An ethics statement has been included to discourage the direct application of LLM content for review-related tasks.}
    \label{fig:gradio}
\end{suppfigure}

\clearpage 
\newpage

\clearpage
\newpage

\clearpage
\newpage

\begin{suppfigure}[htb]
\begin{lstlisting}
Your task now is to draft a high-quality review outline for a Nature family journal for a submission titled <Title>: 

```
<Paper_content>
```


======
Your task: 
Compose a high-quality peer review of a paper submitted to a Nature family journal.

Start by "Review outline:".
And then: 
"1. Significance and novelty"
"2. Potential reasons for acceptance"
"3. Potential reasons for rejection", List multiple key reasons. For each key reason, use **>=2 sub bullet points** to further clarify and support your arguments in painstaking details. Be as specific and detailed as possible. 
"4. Suggestions for improvement", List multiple key suggestions. Be as specific and detailed as possible. 

Be thoughtful and constructive. Write Outlines only. 
\end{lstlisting}
\caption{
Prompt template employed with GPT-4 for generating scientific feedback on papers from the \emph{Nature} journal family dataset. 
<Paper\_content> denotes text extracted from the paper, including the paper's abstract, figure and table captions, and other main text sections. 
For clarity and succinctness, GPT-4 was directed to formulate a structured outline of scientific feedback. 
GPT-4 was requested to generate four feedback sections: significance and novelty, potential reasons for acceptance, potential reasons for rejection, and suggestions for improvement.
The feedback was generated by GPT-4 in a single pass.
}
\label{fig:Nature-prompt-generate}
\end{suppfigure}

\clearpage 
\newpage

\begin{suppfigure}[htb]
\begin{lstlisting}
Your goal is to identify the key concerns raised in the review, focusing only on potential reasons for rejection.

Please provide your analysis in JSON format, including a concise summary, and the exact wording from the review.

Submission Title: <Title>

=====Review:
```
<Review_Text>
```

=====
Example JSON format:
{{
    "1": {{"summary": "<your concise summary>", "verbatim": "<concise, copy the exact wording in the review>"}},
    "2": ...
}}

Analyze the review and provide the key concerns in the format specified above. Ignore minor issues like typos and clarifications. Output only json. 
\end{lstlisting}
\caption{
Prompt template employed with GPT-4 for extractive text summarization of comments in LLM and human feedback. The output is structured in JSON (JavaScript Object Notation) format, where each JSON key assigns an ID to a specific point, and the corresponding value provides the content of the point. 
}
\label{fig:extract-comment-prompt}
\end{suppfigure}

\clearpage 
\newpage 

\begin{suppfigure}[htb]
\begin{lstlisting}
Your task is to carefully analyze and accurately match the key concerns raised in two reviews, ensuring a strong correspondence between the matched points. Examine the verbatim closely.

=====Review A:
```
<JSON extracted comments for Review A from previous step>
```

=====Review B:
```
<JSON extracted comments for Review B from previous step>
```

Please follow the example JSON format below for matching points. For instance, if point 1 from review A is nearly identical to point 2 from review B, it should look like this:

{{
"A1-B2": {{"rationale": "<explain why A1 and B2 are nearly identical>", "similarity": "<5-10, only an integer>"}},
...
}}


Note that you should only match points with a significant degree of similarity in their concerns. Refrain from matching points with only superficial similarities or weak connections. For each matched pair, rate the similarity on a scale of 5-10.

5. Somewhat Related: Points address similar themes but from different angles.
6. Moderately Related: Points share a common theme but with different perspectives or suggestions.
7. Strongly Related: Points are largely aligned but differ in some details or nuances.
8. Very Strongly Related: Points offer similar suggestions or concerns, with slight differences.
9. Almost Identical: Points are nearly the same, with minor differences in wording or presentation.
10. Identical: Points are exactly the same in terms of concerns, suggestions, or praises.


If no match is found, output an empty JSON object. Provide your output as JSON only.
\end{lstlisting}
\caption{
Prompt template employed with GPT-4 for semantic text matching to match the points of shared comments between two feedback. 
The input comprises two lists of comments in JSON format obtained from the preceding step. 
GPT-4 was then directed to identify common points between the two lists, and generate a new JSON where each key corresponds to a pair of matching point IDs, and the associated value provides the rationale for the match. 
}
\label{fig:match-comment-prompt}
\end{suppfigure}

\clearpage

\end{document}